\documentclass[letterpaper]{article} 
\usepackage{aaai25}  
\usepackage{times}  
\usepackage{helvet}  
\usepackage{courier}  
\usepackage[hyphens]{url}  
\usepackage{graphicx} 
\usepackage{natbib}  
\usepackage{caption} 
\usepackage{algorithm}
\usepackage{algorithmic}

\usepackage{booktabs}       
\usepackage{amsmath}
\usepackage{amssymb}
\usepackage{mathtools}
\usepackage{amsthm}
\usepackage{graphicx} 
\usepackage{cancel} 
\usepackage{multirow}
\usepackage{subcaption}
\usepackage[capitalise]{cleveref}
\usepackage{makecell}
\usepackage{newfloat}
\usepackage{listings}
\DeclareCaptionStyle{ruled}{labelfont=normalfont,labelsep=colon,strut=off} 
\floatstyle{ruled}
\newfloat{listing}{tb}{lst}{}
\floatname{listing}{Listing}
\title{Rolling Ahead Diffusion for Traffic Scene Simulation}
\author{
  Yunpeng Liu\textsuperscript{\rm 1,\rm 2} 
  Matthew Niedoba\textsuperscript{\rm 1,\rm 2} 
  William Harvey\textsuperscript{\rm 1} 
  Adam \'Scibior\textsuperscript{\rm 2}\\
  Berend Zwartsenberg\textsuperscript{\rm 2}
  Frank Wood\textsuperscript{\rm 1,\rm 2,\rm 3}  
}
\affiliations{
    \textsuperscript{\rm 1}University of British Columbia,
    \textsuperscript{\rm 2}Inverted AI,
    \textsuperscript{\rm 3}Amii

}

\usepackage{bibentry}

\def\method{\text{RoAD}}

\begin{document}

\maketitle

\begin{abstract}
Realistic driving simulation requires that NPCs not only mimic natural driving behaviors but also react to the behavior of other simulated agents. Recent developments in diffusion-based scenario generation focus on creating diverse and realistic traffic scenarios by jointly modelling the motion of all the agents in the scene. However, these traffic scenarios do not react when the motion of agents deviates from their modelled trajectories. For example, the ego-agent can be controlled by a stand along motion planner. To produce reactive scenarios with joint scenario models, the model must regenerate the scenario at each timestep based on new observations in a Model Predictive Control (MPC) fashion. Although reactive, this method is time-consuming, as one complete possible future for all NPCs is generated per simulation step. Alternatively, one can utilize an autoregressive model (AR) to predict only the immediate next-step future for all NPCs. Although faster, this method lacks the capability for advanced planning. We present a rolling diffusion based traffic scene generation model which mixes the benefits of both methods by predicting the next step future and simultaneously predicting partially noised further future steps at the same time. We show that such model is efficient compared to diffusion model based AR, achieving a beneficial compromise between reactivity and computational efficiency. 
\end{abstract}

%

\section{Introduction}
Traffic simulation is essential in the development of autonomous driving systems, particularly for addressing sim-to-real issues during real-world deployment. A key component is the realistic behavior of simulated surrounding agents or non-playable characters (NPCs), as noted by~\cite{gulino2024waymax}. One approach involves replaying recorded driving behaviors, obtained either from overhead drones~\cite{interactiondataset} or onboard vehicle recordings~\cite{sun2020scalability}. According to~\cite{gulino2024waymax}, reinforcement learning policies trained on such driving logs outperform those trained with NPCs controlled by the rule-based reactive planner like the Intelligent Driver Model (IDM)~\cite{treiber2000congested}. However, this log-replay approach has two drawbacks, the collection of driving records is both costly and time-consuming~\cite{rempe2022strive,liu2023video}. In addition, real-world agents react to the ego agent’s behavior with diverse actions, a dynamic not captured by IDM.


An increasingly popular alternative involves controlling simulated NPCs with generative driving behavior models~\cite{suo2021trafficsim, scibior2021imagining, xu2023bits}, which learn to generate realistic driving trajectories from recorded behaviors. These models facilitate the generation of a diverse set of synthetic driving logs. More recently, diffusion model based traffic scene prediction models have become more popular~\cite{guo2023scenedm, jiang2023motiondiffuser, niedoba2024diffusion}. These models generate joint future trajectories for all agents based on initial observations and allow for flexible conditioning at test time with classifier guidance for tasks like adversarial scenario generation~\cite{zhong2023guided} and scenario editing~\cite{niedoba2024diffusion}.

While these diffusion based models are suitable for open-loop simulations, employing them in closed-loop simulations is time-consuming due to the inherently slow generation times of diffusion models. In such setups, one would need to  replan at each simulation timestep to maintain maximum reactivity to the uncontrolled ego agent. Current methods utilizing diffusion models for closed-loop traffic scene planning typically generate a small window of steps ahead, then replan~\cite{chang2023controllable} to create adversarial scenarios. However, this approach may be inefficient, our method improves upon this by partially denoising the future plan  and only predicting a clean subsequent state after realizing all agents' states at the last simulation step.
\raggedbottom


We propose a rolling diffusion-based traffic simulation planner that efficiently plans for all agents in the scene in an autoregressive manner. With a 2-second look-ahead planning window, our model requires four times fewer function evaluations than a typical autoregressive diffusion model for generating 4-second scenarios, while remaining reactive to other agents.
Since the denoising operation cannot benefit from parallel computing, this iterative process becomes the bottleneck in simulation speed. Furthermore, our experiments, which involve training on real-world driving logs~\cite{interactiondataset}, demonstrate that our rolling-ahead traffic planner produces more realistic scenes than our diffusion-based AR baseline, as measured qualitatively and by scene-level displacement metrics.

\section{Background}
\subsection{Diffusion Models}

\begin{figure*}[th]
     \centering
        \begin{subfigure}[b]{0.22\textwidth}
        \centering
\includegraphics[width=1\linewidth,keepaspectratio]{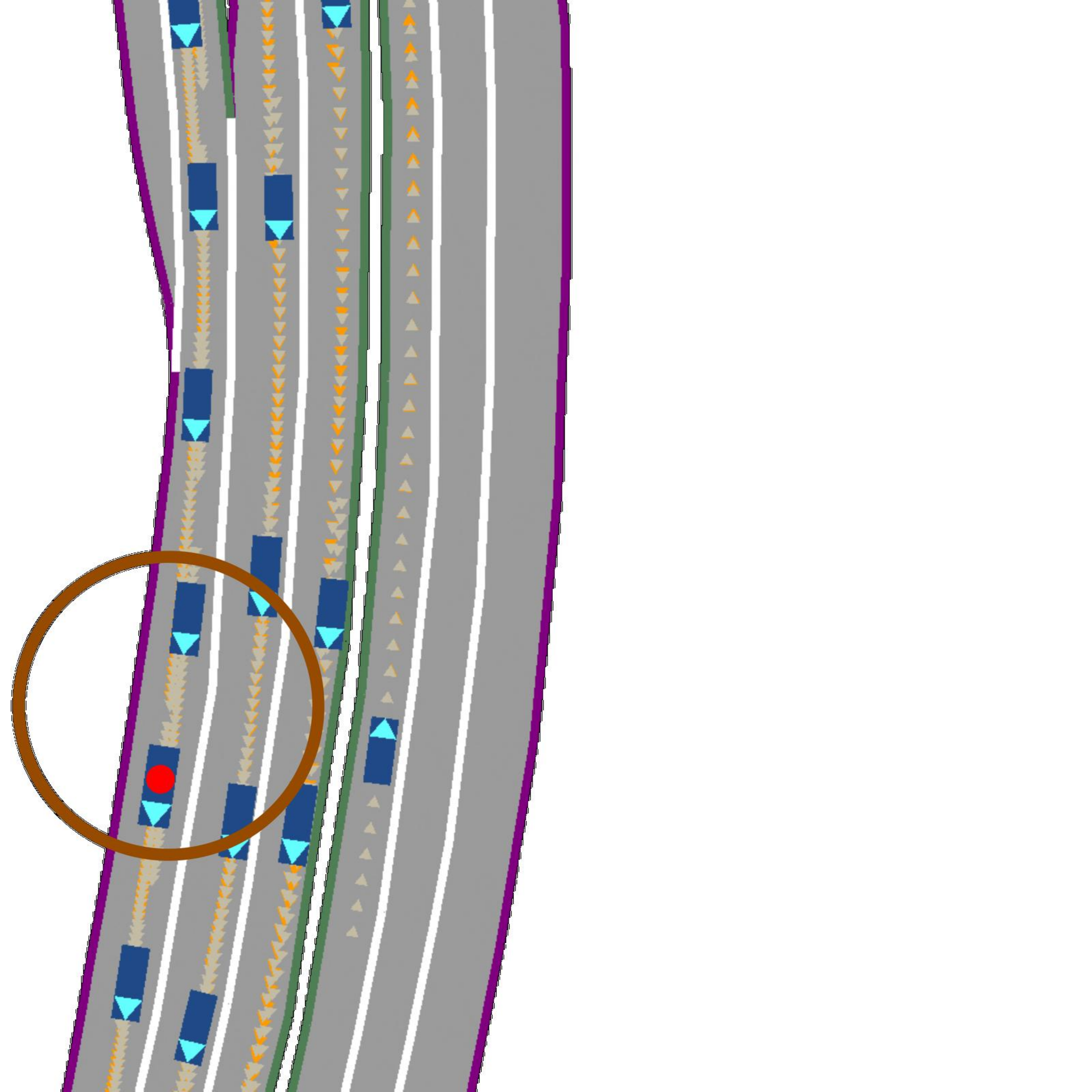}
    \end{subfigure}
            \begin{subfigure}[b]{0.22\textwidth}
        \centering
\includegraphics[width=1\linewidth,keepaspectratio]{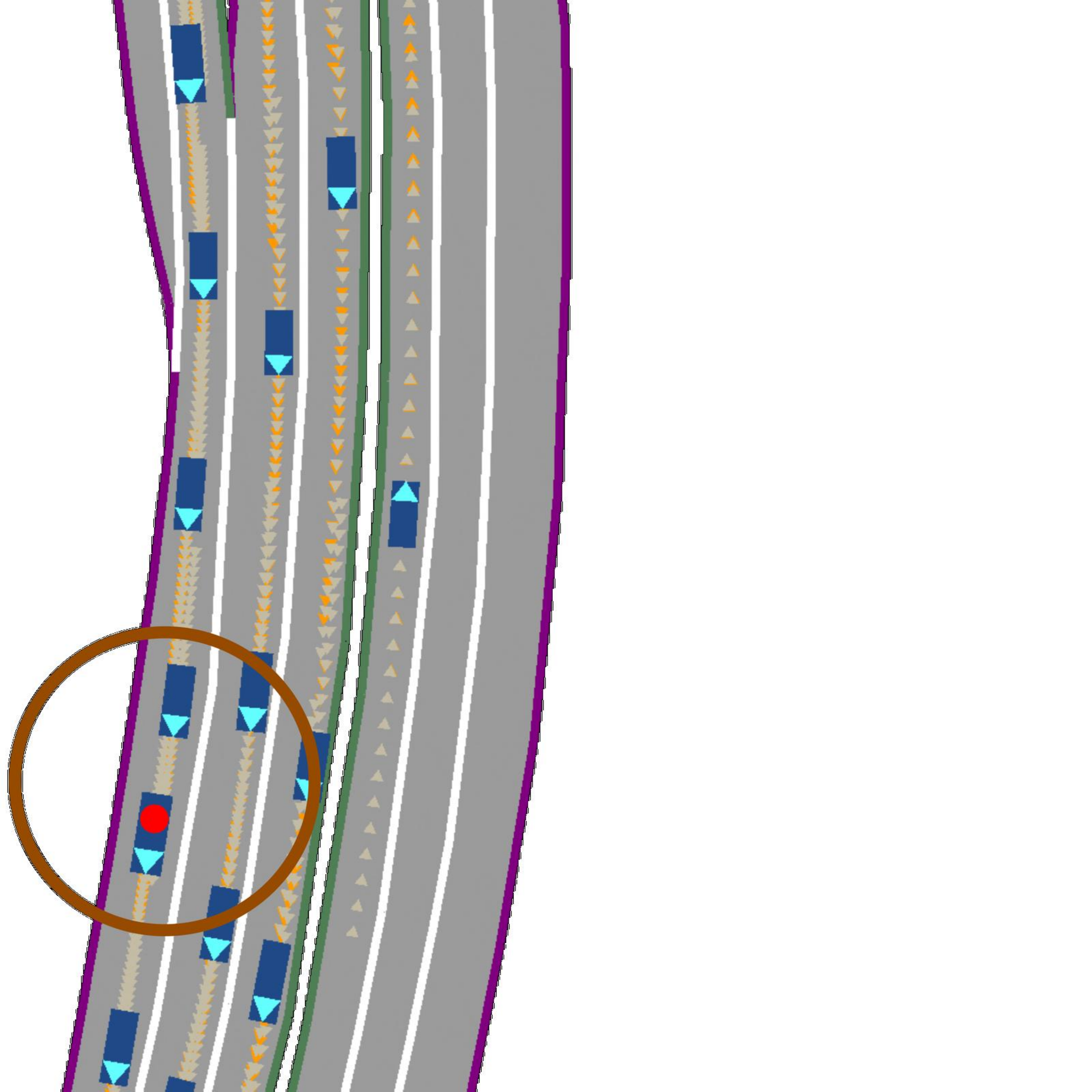}
    \end{subfigure}
        \begin{subfigure}[b]{0.22\textwidth}
        \centering        \includegraphics[width=1\linewidth,keepaspectratio]{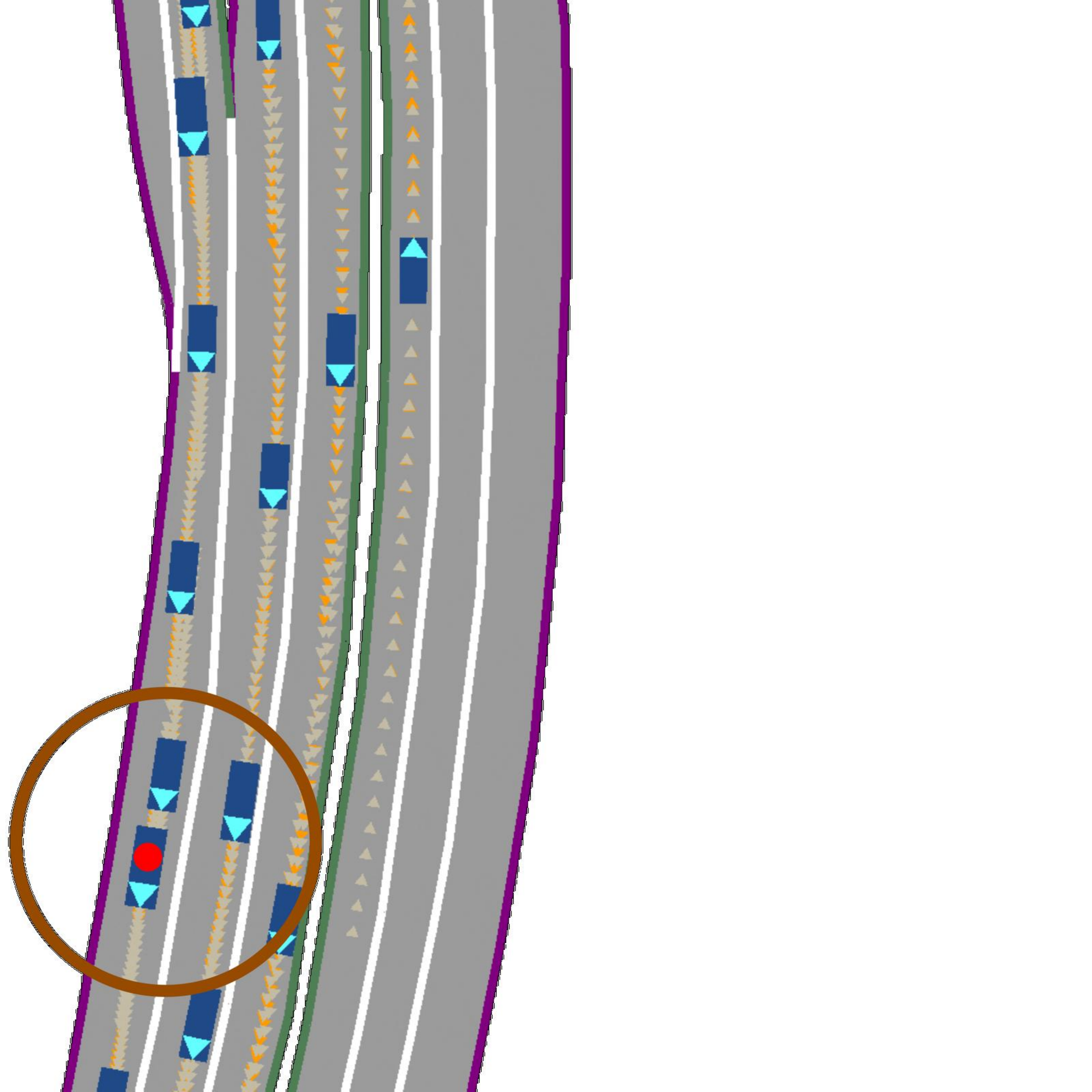}
    \end{subfigure}
        \begin{subfigure}[b]{0.22\textwidth}
        \centering        \includegraphics[width=1\linewidth,keepaspectratio]{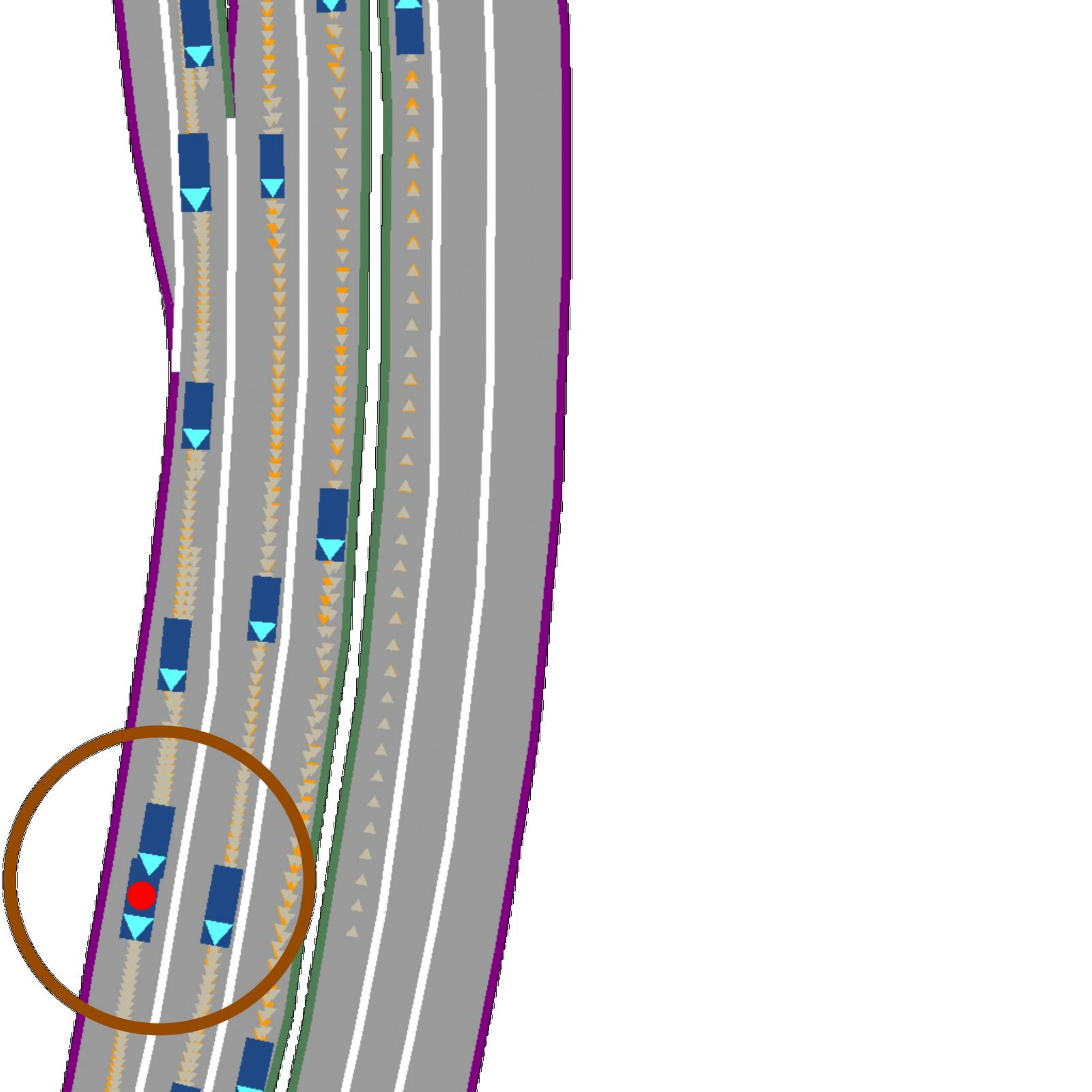}
            \end{subfigure}
    
     \begin{subfigure}[b]{0.22\textwidth}
        \centering        \includegraphics[width=1\linewidth,keepaspectratio]{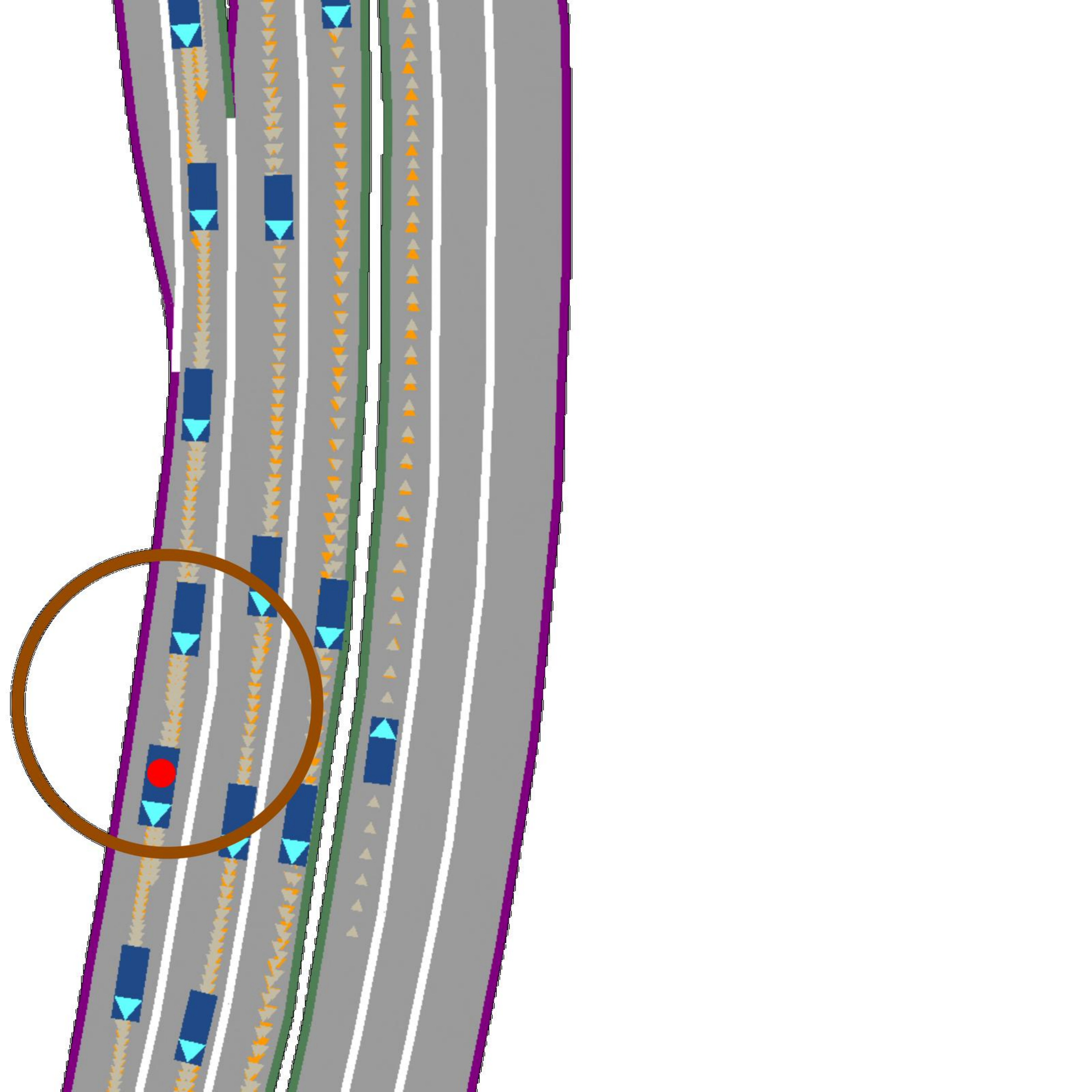}
    \end{subfigure}
    \begin{subfigure}[b]{0.22\textwidth}
        \centering        \includegraphics[width=1\linewidth,keepaspectratio]{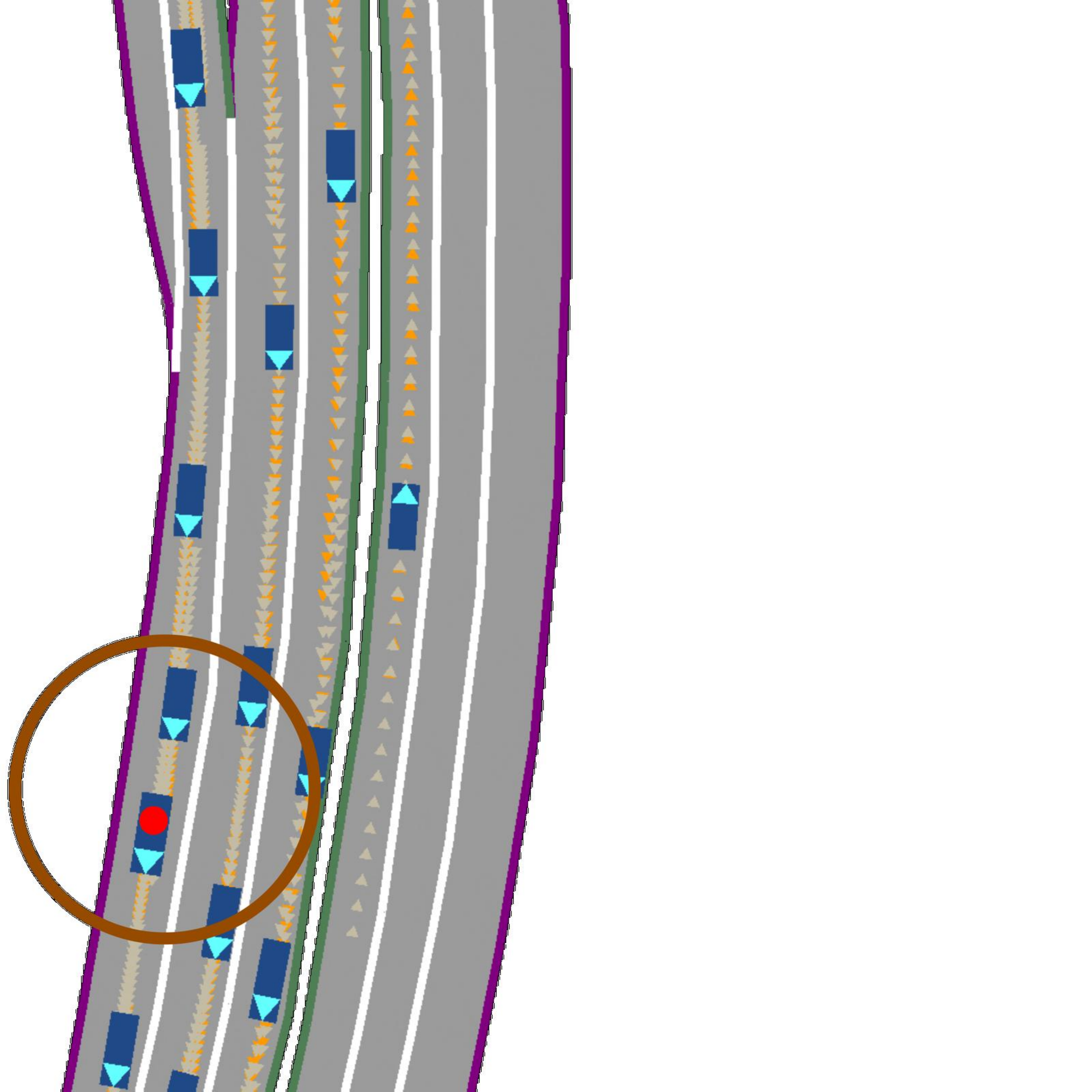}
    \end{subfigure}
    \begin{subfigure}[b]{0.22\textwidth}
        \centering        \includegraphics[width=1\linewidth,keepaspectratio]{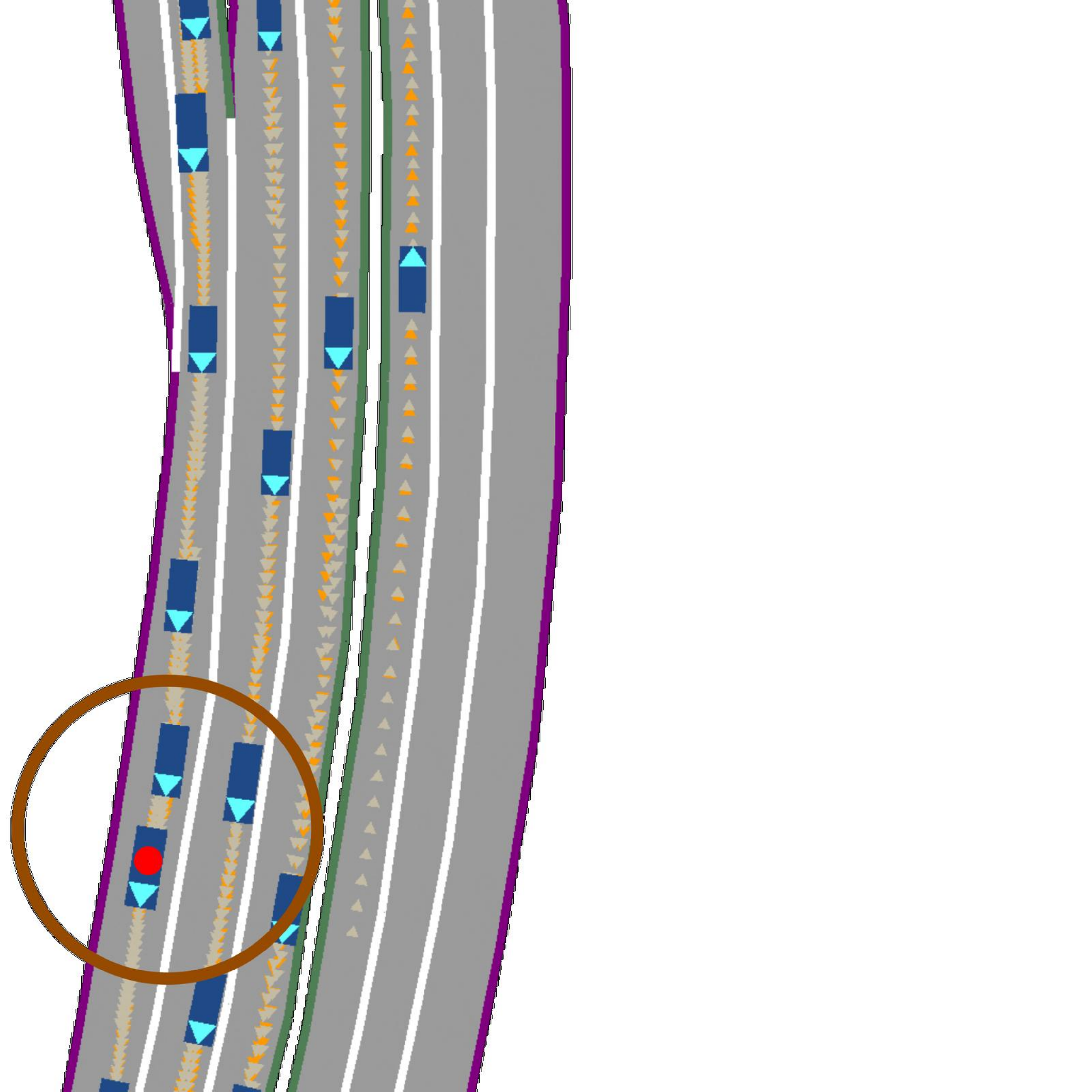}
    \end{subfigure}
        \begin{subfigure}[b]{0.22\textwidth}
        \centering        \includegraphics[width=1\linewidth,keepaspectratio]{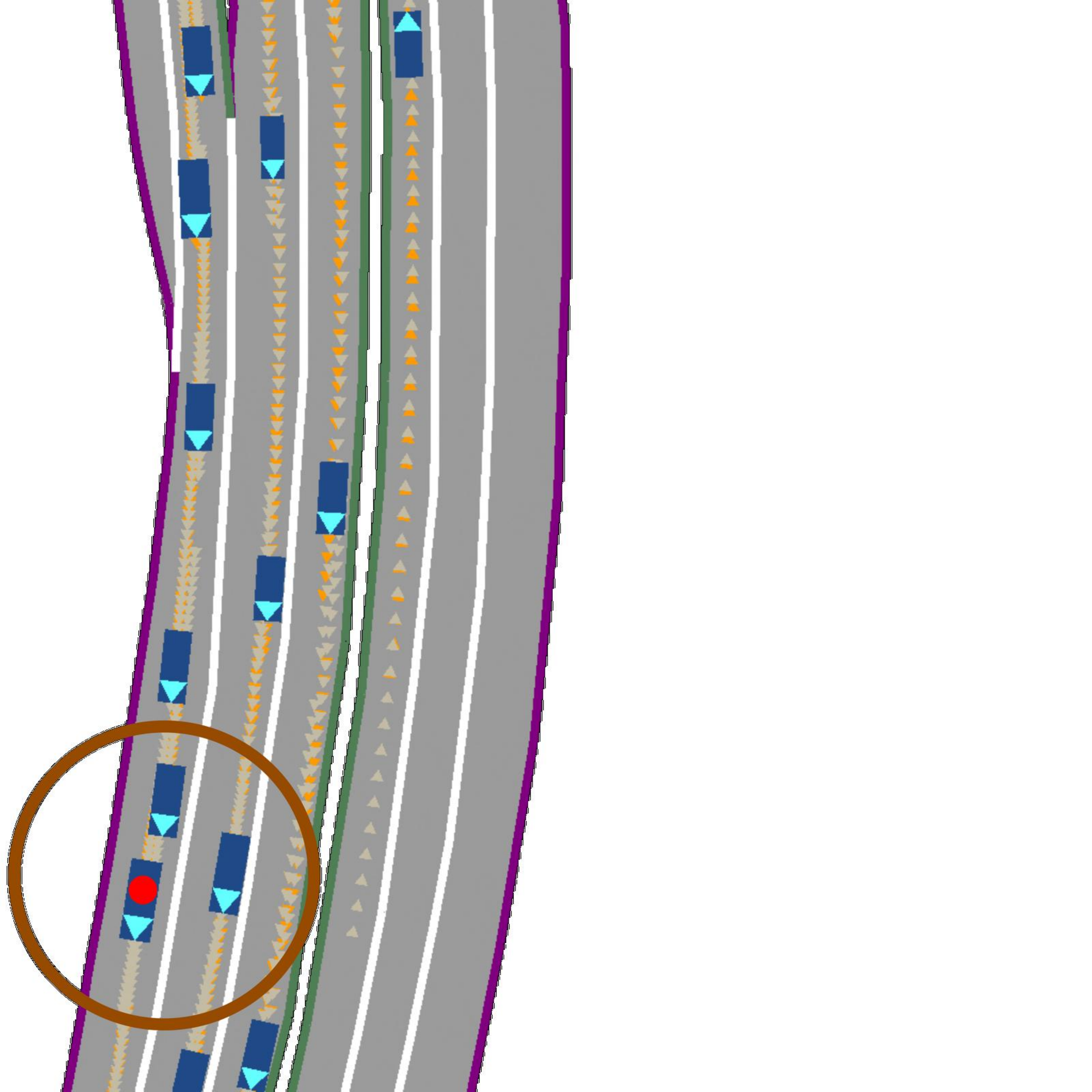}
    \end{subfigure}
     \begin{subfigure}[b]{0.22\textwidth}
        \centering        \includegraphics[width=1\linewidth,keepaspectratio]{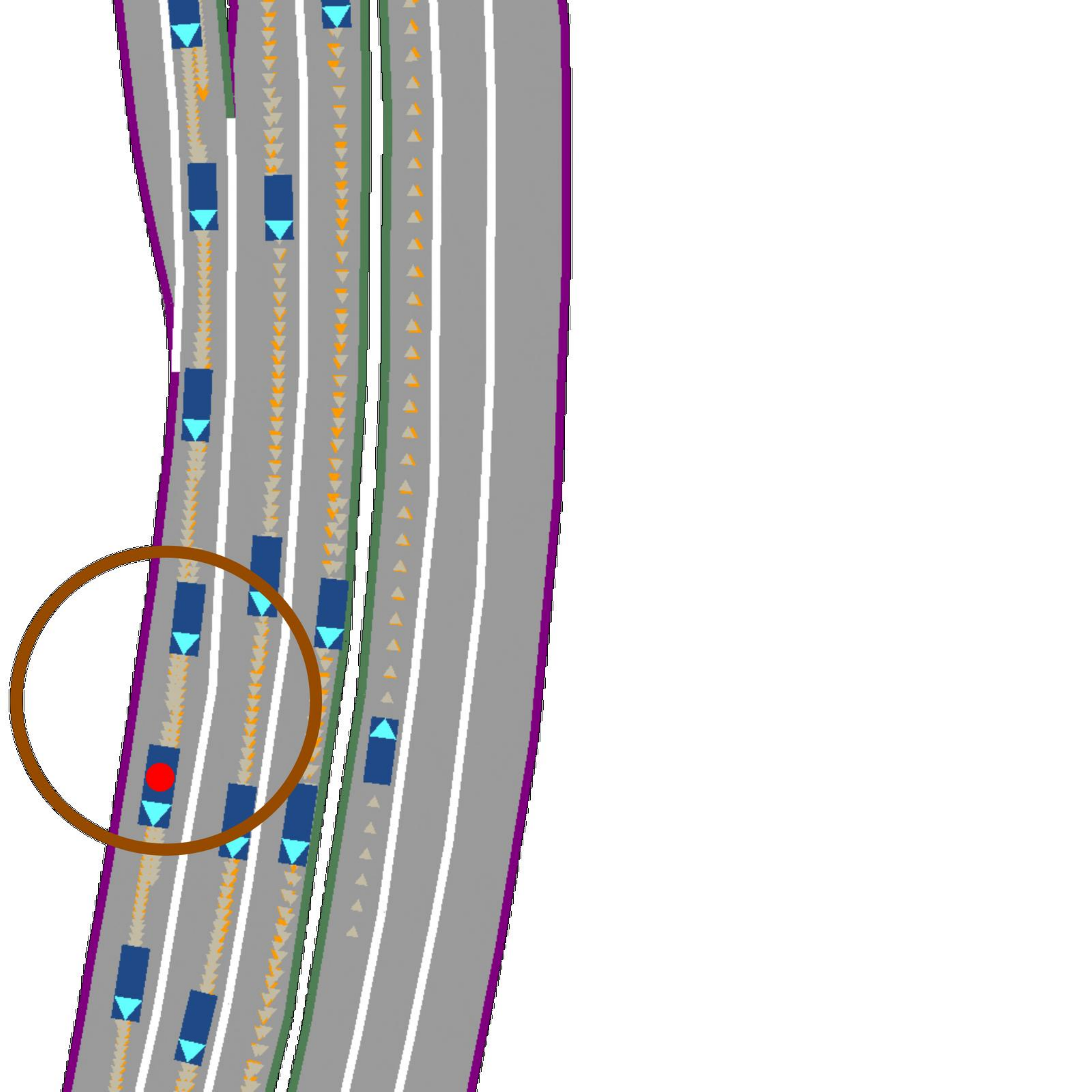}
    \end{subfigure}
    \begin{subfigure}[b]{0.22\textwidth}
        \centering        \includegraphics[width=1\linewidth,keepaspectratio]{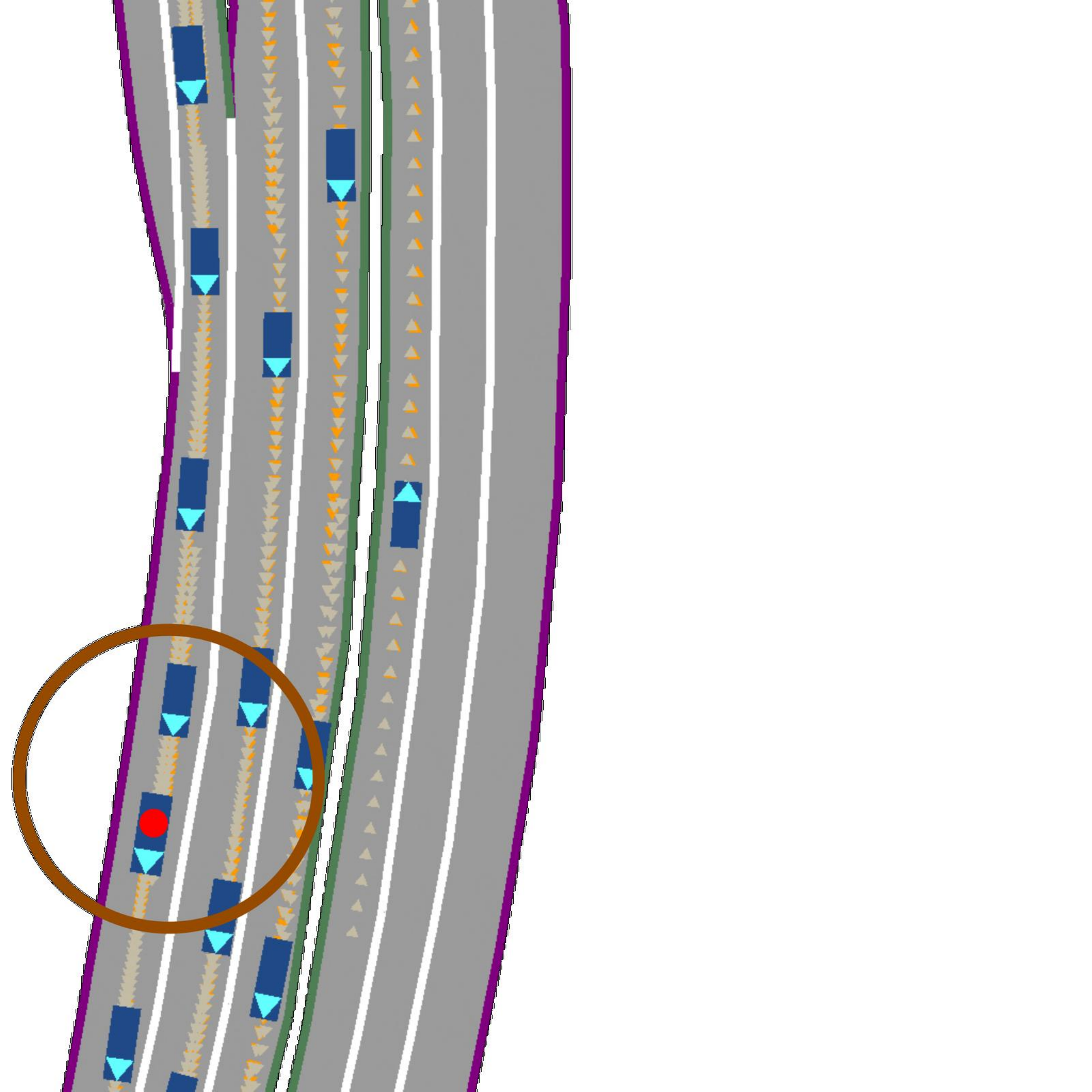}
    \end{subfigure}
    \begin{subfigure}[b]{0.22\textwidth}
        \centering        \includegraphics[width=1\linewidth,keepaspectratio]{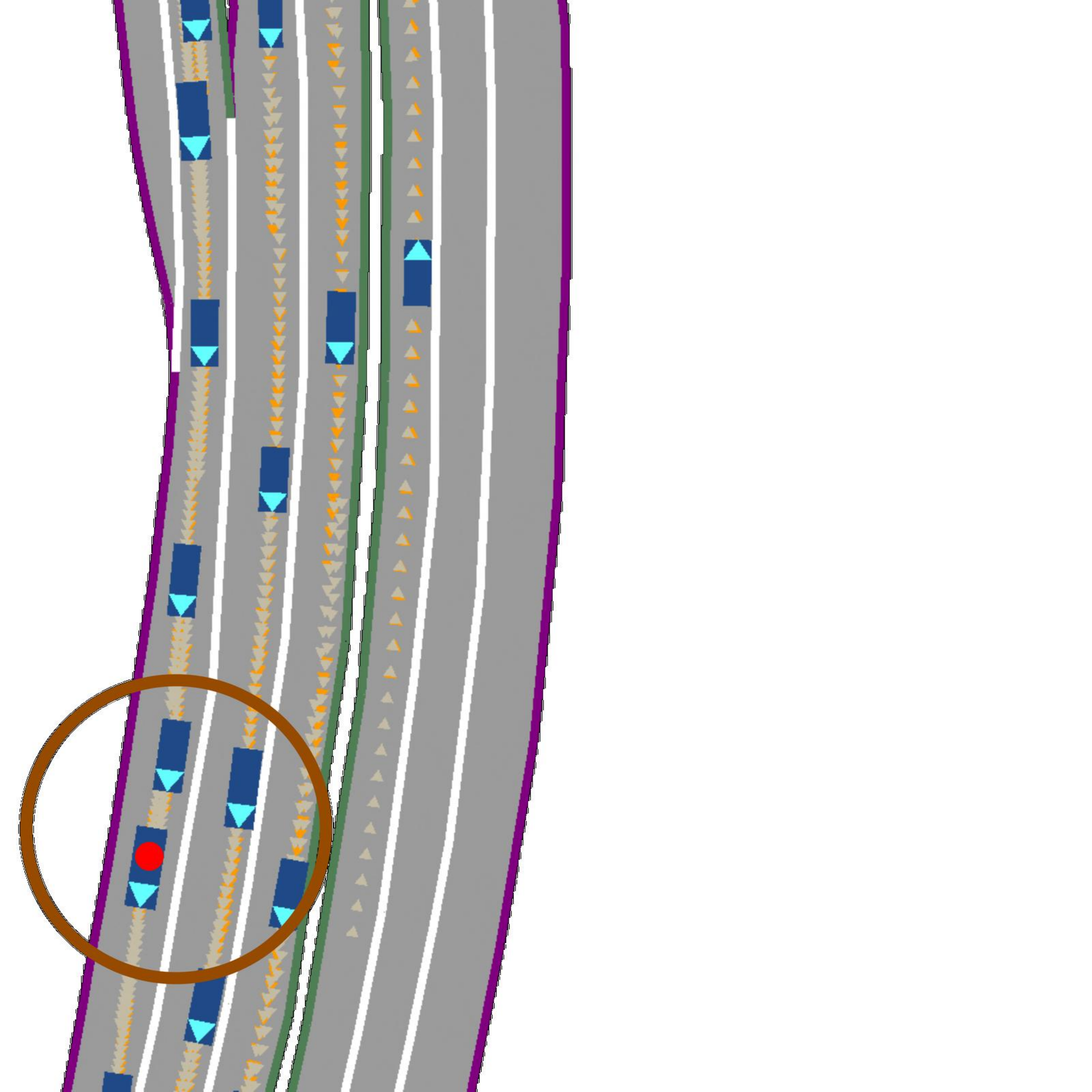}
    \end{subfigure}
        \begin{subfigure}[b]{0.22\textwidth}
        \centering        \includegraphics[width=1\linewidth,keepaspectratio]{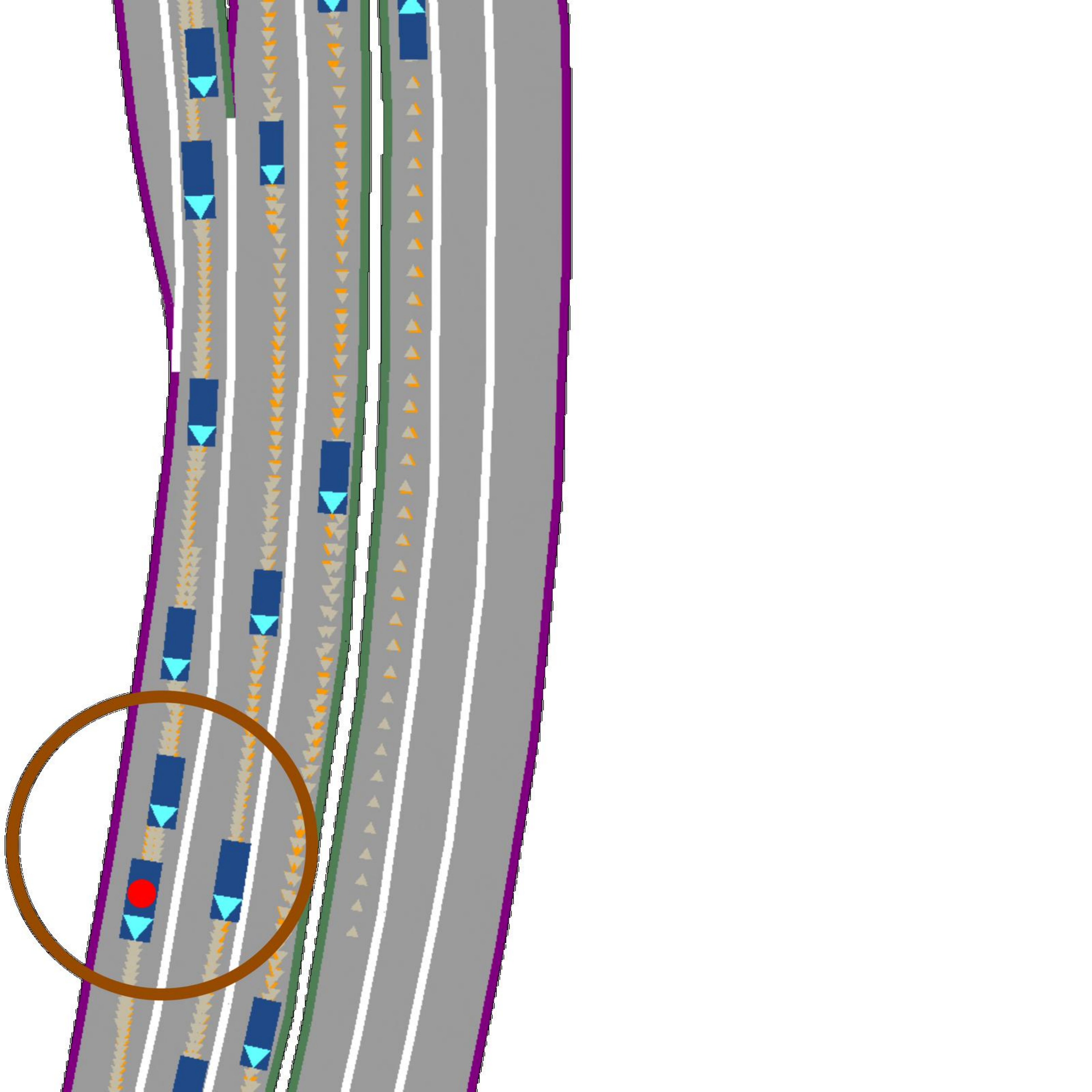}
    \end{subfigure}
    \caption{From top to bottom row: DJINN~\cite{niedoba2024diffusion}, autoregressive (AR), and \method{} (Ours). The adversarial agent, marked with a red dot, follows its replay log and slows down to reach only half its trajectory by the end of the simulation. Brown circles highlight the interaction region. The agents controlled by the \method{} and AR models slow down to react to the adversarial agent, while agents controlled by the DJINN model do not. Ground truth trajectories are shown in gray, and predicted trajectories are shown in orange.}
    \label{fig:traj-v-viz}
\end{figure*}

Diffusion models~\cite{sohl2015deep,ho2020denoising,kingma2021variational,karras2022elucidating,song2020score} are probabilistic generative models which have recently been applied to the problem of traffic scenario modelling. They are defined based on a forward process which gradually adds Gaussian noise to data.
The amount of noise added  at any point is based on the ``time'' in the forward process, $\tau$. We will refer to the copy of the data at a certain diffusion timestep as $\boldsymbol{x}_\tau$. When $\tau$ equals zero, $\boldsymbol{x}_\tau = \boldsymbol{x}_0$ is the original data with no noise. For positive $\tau$, the distribution of $\boldsymbol{x}_\tau$  given $\boldsymbol{x}$ is
\begin{equation}
    q(\boldsymbol{x}_{\tau}|\boldsymbol{x}_0) = \mathcal{N}(\boldsymbol{x}_{\tau};  \alpha_{\tau}\boldsymbol{x}_0, \sigma_{\tau}^{2}\boldsymbol{I}), \label{eq:diff_forward}
\end{equation}
where $\alpha_{\tau}$ and $\sigma_{\tau}$ are $\tau$-dependent scalars and $\tau \in{[0,1]}$. In this paper, we will set $\alpha_\tau = 1$ for all $\tau$ following EDM~\cite{karras2022elucidating}. We will define $\sigma_\tau$ to be an increasing function which is zero when $\tau$ is zero and large when $\tau=1$. The signal-to-noise ratio is defined by 
\begin{equation}
    \text{SNR}(\tau) = \frac{\alpha_{\tau}^2}{\sigma_{\tau}^2}.
\end{equation}
Given our choices of $\alpha_\tau$ and $\sigma_\tau$, the signal to noise ratio is small for $\tau=1$. This means that $q(\boldsymbol{x}_1|\boldsymbol{x}_0)$ will be well-approximated by a zero-mean Gaussian, and therefore the marginal $q(\boldsymbol{x}_1|\boldsymbol{x}_0) = \int q(\boldsymbol{x}_1|\boldsymbol{x}_0) p_\text{data}(\boldsymbol{x}_0) \mathrm{d}\boldsymbol{x}_0$ will also be roughly Gaussian.


It is possible to "invert" this forward process, yielding a reverse process that can transport samples from $q(\boldsymbol{x}_1)$ to $q(\boldsymbol{x}_0)$, as shown by \cite{song2020score}. This reverse process maps a roughly-Gaussian distribution to the data distribution, allowing us to make realistic samples out of noise. Doing so only requires an approximation of the score function $\nabla_{\boldsymbol{x}_\tau} \log q(\boldsymbol{x}_\tau)$. Such an approximation can be learned by optimizing the loss~\cite{song2020score}
\begin{equation} \label{eq:diffusion-objective}
    \begin{split}
            &\mathcal{L}_\text{diffusion}(\theta) =\\
&\mathbb{E}_{p_\text{data}(\boldsymbol{x}_0) q(\boldsymbol{x}_{\tau}|\boldsymbol{x}_{0})u(\tau)} \left[ \omega(\tau)  \|D_{\theta}(\boldsymbol{x}_{\tau}; \tau) - \boldsymbol{x}_0 \|_2^2 \right]
    \end{split}
\end{equation}
where $u(\tau)$ is a distribution over diffusion times to use during training, $\omega(\boldsymbol{\tau})$ is a function that weights the contribution of each timestep to the loss, and $D_\theta(\cdot,\cdot)$ represents a neural network that produces an output of the same shape as $\boldsymbol{x}_0$. This neural network learns to estimate clean data given noisy data, and such an estimate can be used to produce an estimate of the score function~\cite{song2020score}. This can be used to simulate the SDE that produces sampled clean data $\boldsymbol{x}_0$ given samples from a Gaussian approximating $q(\boldsymbol{x}_1)$. Through out the paper we use $\tau$ as the time in the diffusion process and $t$ as the chronological time within a scenario. Note that the diffusion model can be made conditional on any extra information by inputting the extra information into the neural network in \cref{eq:diffusion-objective} and providing it to the neural network in the same way at test-time~\cite{tashiro2021csdi}.

\subsection{Temporally Correlated Diffusion Models}



The above formulation of diffusion models becomes  less efficient in terms of memory and computational resources as the dimensions of $\boldsymbol{x_0}$ grow, particularly for long sequence modeling. The authors of the rolling diffusion model (RDM)~\cite{ruhe2024rolling} introduce a method by modelling a sliding window of $\boldsymbol{x_0}$, given the assumptions that elements that are far in the past from the sliding window are irrelevant.The authors propose assigning temporally correlated noise levels to the elements within the sliding window, introducing a temporal inductive bias to the model.
Previous work has shown that this particular inductive bias helps generating diverse and high quality samples in long term human motion predictions~\cite{zhang2023tedi} and being efficient in NLP tasks like summarization and translation~\cite{wu2024ar}. RDM formalizes the temporal noise correlation approach and provides a local and global perspective of modelling long sequences of videos.



\begin{figure*}[t]
\centering
\includegraphics[width=0.7\textwidth]{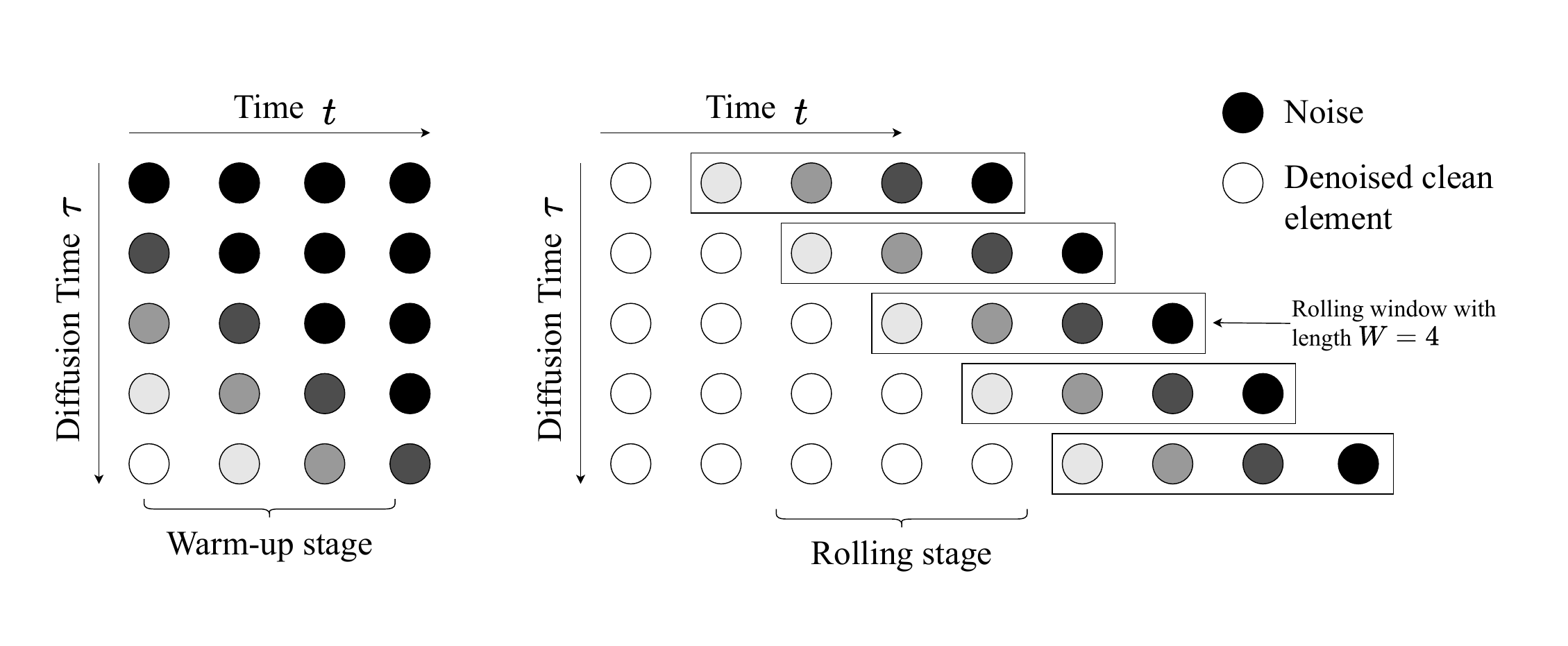} 
\caption{Rolling Diffusion Model. Columns represent sequence timesteps and rows represent diffusion timesteps.  Circles are shown in white if the corresponding sequence timestep is fully denoised; black if the sequence timestep is pure noise; and grey if in between. During the denoising process, the SNR for each element in the rolling window depends on the local diffusion time $\tau_w$ which can be calculated using ~\cref{eq:taww_warm} or~\cref{eq:taww_rolling}, depending on whether it is in the warm-up or rolling stage.}
\label{fig:rdm}
\end{figure*}

We begin our discussion by examining a general rule for diffusion models that incorporate temporal noise correlation for sequences. We then delve into the specific mechanisms of the RDM, reviewing both its forward and reverse processes along with its objective function, and the two operating stage as depicted in Figure~\ref{fig:rdm}.

Given a long sequence of data with length $T$, RDM approaches the problem from a local perspective by examining a single sliding window of length $W$. Within this framework, RDM defines a function $g$ that maps a global diffusion step $\tau$ to a local diffusion step $\tau_w \in [0,1]$ given the local window index $w$. The fundamental rule for diffusion models with temporal noise correlation is
\[
\text{SNR}(\tau_{w+1}) < \text{SNR}(\tau_w),
\]
indicating  the temporal nature of the sequence results in increased uncertainty as time progresses. As shown in Figure~\ref{fig:rdm}, darker-colored circles represent higher uncertainty because they are at a higher noise level.

Given a sampled sequence index $t$, The forward process within the local window in RDM is
\begin{equation}
 q( \boldsymbol{x}^{t:t+W}_{\tau} | \boldsymbol{x}^{t:t+W}_{0}) = \prod^{t+W-1}_{w=t} \mathcal{N}(\boldsymbol{x}^{w}_{\tau};  \alpha_{\tau_w}\boldsymbol{x}_0^{w}, \sigma_{\tau_w}^{2}\boldsymbol{I}),
 \label{eq:rdm_forward}
\end{equation}
where the diffusion parameter $\alpha_{\tau_w}$ and $\sigma_{\tau_w}$ are local diffusion step $\tau_w$-dependent scalars rather than $\tau$ in ~\cref{eq:diff_forward}. This represents the fact that more noise is added to the later frame as $\alpha_{\tau}$ and $\sigma_{\tau}$ is monotonically increasing with respect to $\tau$. 

For the reverse process, RDM defines
\begin{align}
&p_{\theta}(\boldsymbol{x}^{t:t+W}_{\tau-1}|\boldsymbol{x}^{t:t+W}_{\tau}) 
:= \nonumber\\&\prod_{w=t}^{t+W-1} q(\boldsymbol{x}_{\tau-1}^w|\boldsymbol{x}_{\tau}^{t:t+W}, \quad \boldsymbol{x}^w = f_{\theta}(\boldsymbol{x}_{\tau}, \tau_w)).
\end{align}
The benefit arising from such formulation is that instead of training on the full sequence $T$ which is memory in-efficient and complex, RDM's objective function is defined only within a sampled sliding window with length $W$,


\begin{align}
\mathbb{E}_{\substack{\tau \sim \mathcal{U}(0,1), \\ \boldsymbol{x}^{t:t+W}_{\tau} \sim  
q}} 
\bigg[ 
\sum_{w=t}^{t+W-1} \omega(\tau_w)
\| D_\theta(\boldsymbol{x}^{t:t+W}_{\tau}; \tau_w) 
- \boldsymbol{x}^w_0 \|_2^2 
\bigg].
\label{eq:rdm_obj}
\end{align}

There are two stages of RDM, the warm-up stage and the rolling stage. In the warm-up stage, the model handles the initial boundary condition by generating from white noise, as shown in the first row of Figure~\ref{fig:rdm} (left), and denoises it to produce one clean element  and partially denoised future elements in the sliding window as shown in the bottom row. Once it reaches the temporal correlated noise stage (Bottom row of Figure~\ref{fig:rdm} left), RDM takes few denoising steps for the next step prediction shown in Figure \ref{fig:rdm} (right).  This requires the model to train two tasks, where $\beta$ controls the training task distribution and for each tasks, RDM designs an associated function $g$ for calculating the local diffusion time $\tau_w$ given $\tau$ and window index $w$. In addition, we can condition $n$ number of clean observations within the sliding window, $g$ is defined for warm-up and rolling stage as 
\begin{align}
    &g_{\text{warm-up}} (\tau, w) := \max(\min(\frac{w}{W}+\tau, 1.0), 0.0) \label{eq:taww_warm} \\ 
    &g_{\text{rolling}} (\tau, w) := \max(\min(\frac{w+\tau-n}{W-n}, 1.0), 0.0) \label{eq:taww_rolling},
\end{align}
where $n$,$W$ are application-dependent hyperparameters.

\section{Related Work}

\subsection{Traffic Simulation with Diffusion Models}

Predicting the motion of road users is a critical task for autonomous vehicle driving or simulation. For this reason, the number of methods which have attempted to model traffic behavior is vast. The literature contains a variety of techniques for modelling the distribution of driving behavior, including mixture models \cite{ChaiSBA19, CuiRCLNHSD19, NayakantiAZGRS23}, variational autoencoders \cite{scibior2021imagining,suo2021trafficsim}, and generative adversarial networks \cite{zhao2019multi}. 

Our work builds upon recent methods which model driving behavior using diffusion models. In CTG \cite{zhong2023guided}, the authors model the motion of each agent in the scene independently with a Diffuser \cite{JannerDTL22} based diffusion model. The authors of \cite{chang2023controllable} also model agent motions via diffusion, with a focus on controllability. By contrast, most other diffusion based traffic models model entire traffic scenes. This includes MotionDiffuser \cite{jiang2023motiondiffuser}, Scenario Diffusion \cite{pronovost2023scenario} and SceneDM \cite{guo2023scenedm} which all diffuse the joint motion of all agents in the scene. Our work builds directly on that of DJINN \cite{niedoba2024diffusion}, which utilizes a transformer based network to generate joint traffic scenarios based on a variable set of agent state observations. Crucially, due to the expensive computational cost of diffusion model sampling, only CTG \cite{zhong2023guided} utilize their model for closed-loop scenario simulation. Twice per second, they incorporate new state observations and resample trajectories for each agent. By comparison, our method does not require iterative replanning, greatly improving simulation speed.


\section{Methods}

\subsection{Problem Formulation}
We refer to the motion of $A$ agents across $T$ discrete times in an environment $\mathcal{M}$ as a traffic scenario. Formally, we define the scenario as $\boldsymbol{x} \in \mathbb{R}^{A \times T \times 3}$, where we represent the state of each agent $a \in A$ at time $t \in T$ as the combination of its 2D position and 1D orientation. We introduce a probabilistic planner $\pi_{sim}$ which jointly predicts the future states for all agents, conditioned on static map information $\mathcal{M}$ and previously observed agent states $x_{obs} \in \mathbb{R}^{A \times t_{obs} \times 3}$ .

A more difficult form of this planning problem is closed-loop traffic simulation. In closed-loop simulation one agent, known as the \textit{ego agent} $a^{ego}$, is typically controlled by a standalone motion planner $\pi_{ego}$ which may be a black-box and which may cause the ego agent to drive very differently to any agents in the training data and thus is not amenable to accurate prediction by the traffic scenario planner $\pi_{sim}$.  At each time step $t$, the standalone motion planner $\pi_{ego}$ plans the single next step for the ego-agent given the entire history of the scenario, $\boldsymbol{x}^{0:t}$ and the map $\mathcal{M}$. The closed-loop traffic simulation problem is to model the behavior of every other agent in this scene, including potential interactions with the ego agent. Since the state of the ego agent is neither controllable nor known in advance, the traffic scenario planner $\pi_{sim}$ must continually update its plan to be continuously conditioned on the most recent state and actions of the ego agent.

\subsection{Replanning with a joint prediction model}
Our baseline planner relies on a conditional diffusion model $p(\boldsymbol{x}^{t_{obs}:T}|\boldsymbol{x}^{0:t_{obs}}, \mathcal{M},\boldsymbol{c})$, which jointly predicts the scenario for all agents in the scene up to time $T$ given the map $\mathcal{M}$ and additional conditioning information $c$. Although diffusing the joint states of all agents is a flexible way of modelling the distribution of traffic scenarios, the model does not respond to the ego agent trajectories which deviate from modelled behavior. To mitigate this, one option is to regenerate the traffic scenario after each simulator step to incorporate new ego agent state observations. We select DJINN~\cite{niedoba2024diffusion} as our conditional diffusion model and we denote this method of iterative planning as DJINN-MPC as it resembles a traditional model predictive control loop.
This allows the scenario simulation planner $\pi_{sim}$ to adjust its predictions at every simulation step in response to the standalone ego agent in the scene. 

\subsection{Diffusion based autoregressive model (Diff-AR)}
One key drawback of DJINN-MPC is that we must fully diffuse a new traffic scenario at every simulator step, at significant cost. As an alternative, one can train an diffusion based autoregressive model as the simulation planner, which factorizes the conditional probability as
\[
p(\boldsymbol{x}^{t_{obs}:T}|\boldsymbol{x}^{0:t_{obs}}, \mathcal{M},\boldsymbol{c}) = \prod_{t=t_{obs}}^{T-1} p(\boldsymbol{x}_t | \boldsymbol{x}^{0:t-1}_{0}, \mathcal{M},\boldsymbol{c}).
\]
Given the past observations, the model only predicts one subsequent step. 
In practice, the history of past observations $\boldsymbol{x}^{0:t-1}_{0}$ is truncated to a fixed length. Compared to the previous method, Diff-AR is slightly more efficient as it only denoises the single-step future from scratch. However,  Diff-AR cannot anticipate other agents' long-term behaviors beyond the immediate next step which is important for effective planning in many traffic scenarios.


\subsection{Rolling ahead autoregressive model}
We propose a rolling diffusion based model~(\method{}) for traffic scenario planning based on RDM. We start by providing an overview of our autoregressive traffic planner, then discuss some details of our design choices on the diffusion process and the model with our updated objective function.

 We utilize a sliding window of length $W$, which is much smaller than the scenario length $T$. This sliding window includes $t_{obs}$ clean observations for all agents. Within this window, only the $t_{obs+1}$th state is fully denoised at each scenario time for all agents, while the remainder of the sequence undergoes partial denoising. At the next simulation step, we then shift the sliding window and repeat the previous process. By focusing on a smaller window and selectively denoising, our approach maintains computational efficiency while preserving the ability to adaptively plan for the immediate future.

We follow the design choices from EDM~\cite{karras2022elucidating} in designing our diffusion process. Given a local diffusion time $\tau_w$, during training, our $\sigma_{\tau}$ is a continuous version of the sampling noise schedule in EDM,
\[
\sigma_{\tau} =  ({\sigma_{\max}}^{\frac{1}{\rho}} + \tau ({\sigma_{\min}}^{\frac{1}{\rho}} -   {\sigma_{\max}}^{\frac{1}{\rho}} ))^{\rho},
\]
where we keep the default hyper-parameter choice for $\sigma_{\max}$, $\sigma_{\min}$ and $\rho$ from~\cite{karras2022elucidating}.
We apply the Heun 2nd order sampler to sample at prediction time with the same hyper-parameter reported in EDM.  We referred the reader to EDM for the detailed denoising algorithms.



As we are interested in modelling a joint traffic planner for all agents in the scene, we diffuse in a global coordinate. We adopt the map representation from~\cite{niedoba2024diffusion} where $\mathcal{M}$ is represented as an unordered set of polylines, each polylines describing lane centers and normalized for length and scale to match the agent states. Our model, built on a transformer-based architecture utilizes a feature tensor shaped \([A, W, F]\) to process agent trajectories and map information. It embeds noisy and observed states, temporal indices, and the local diffusion step $\tau_w$ into high-dimensional vectors with feature dimension $F$. We apply per-agent positional embeddings to these feature vectors, which are then fed into a series of transformer blocks that perform self-attention in the time and agent dimensions, as well as cross-attention with the map features.


Rather than denoise the sequence one by one as in ~\cref{eq:rdm_obj}, our transformer architecture jointly predicts the score for all noisy states in the window. Denote $\boldsymbol{x}^W$ as the sliding window of interest. Our score estimator $D_{\theta}$ takes in $\boldsymbol{x}^W_{\tau}$, $\boldsymbol{\tau}$, and the map $\mathcal{M}$, along with additional conditional information $\boldsymbol{c}$ that includes the dimensions of each agent. Our updated objective function is
\begin{align}
    \mathbb{E}_{\boldsymbol{x}^W_0, \boldsymbol{\tau}, \boldsymbol{x}^W_{\tau}} [\omega(\boldsymbol{\tau})  \|D_{\theta}(\boldsymbol{x}^W_{\tau}, \mathcal{M}, \boldsymbol{c}, \boldsymbol{\tau}) - \boldsymbol{x}^W_0 \|_2^2].
\end{align}
Note that $\boldsymbol{x}^W_{\tau}$ is sampled from RDM forward process defined in ~\cref{eq:rdm_forward} that contains states with noise level according to $\tau_w$. While local diffusion time $\tau_w$ depends on the window index $w$ and the global diffusion steps $\tau$, all agents in the scene has a consistent local diffusion time $\tau_w$. Therefore, our score estimator $D_{\theta}$ takes a vector  $\boldsymbol{\tau} = \{\tau_w\}_{w=0}^{W}$ to reflect the temporal correlation of different nose levels in $\boldsymbol{x}^W_{\tau}$. The weighting term is also a vector that takes  vector $\boldsymbol{\tau}$ as input then assign different weights according to each $\tau_w$. 

While our rolling ahead autoregressive model is efficient for long traffic scenario planning, the partially denoised future plan affects the reactivity of our model.  In traffic simulation, such degradation may cause a higher collision rate with the uncontrolled ego agent in the scene.  The reactivity of the model depends on the $\text{SNR}$ of future states. We empirically evaluate the reactivity of our model compared to the AR baseline in our experiment section.

\subsection{Conditioning Augmentation}

We have empirically found that noise conditioning augmentation, as described by~\cite{ho2022cascaded}, is essential for all models operating in an autoregressive manner. This augmentation is critical for autoregressive human motion generation~\cite{yin2023controllable}, cascaded diffusion models for class-conditional generation, and super-resolution video generation~\cite{ho2022imagen}. Noise conditioning augmentation enhances the model's robustness against generated noise, which serves as observations for subsequent predictions~\cite{ho2022cascaded}, and it mitigates the risk of the model overfitting to its autoregressive nature. In the context of traffic simulation, this augmentation aids in generating smooth trajectories and ensures that the model does not ignore other conditional factors, such as the presence of other agents and, importantly, the map \(M\) of the environment. Previous work on diffusion-based traffic simulation~\cite{zhang2023tedi, chang2023controllable} circumvents the issue of noisy observations by relying on a kinematic model to produce smooth trajectories; however, our autoregressive traffic simulation planner does not require such a kinematic model.

We follow~\cite{ho2022cascaded} to employ conditioning augmentation for our rolling ahead traffic planner with an important modification.  During training, given a sampled training segment \(\boldsymbol{x}^W\) with length \(W\), and \(n\) observations \(\boldsymbol{x}^{obs}\) within this segment, we apply Gaussian noise augmentation to the \(\boldsymbol{x}^{obs}\), where the noise level \(\sigma_{\tau_{ca}}\) is sampled uniformly between $\sigma_{\text{min}}$ and $\sigma_{\text{max}}$. Unlike~\cite{ho2022cascaded}, we found jointly predicting all elements within the sliding window, including the noised observations is enssential for our application. At testing time, we apply Gaussian noise with $\sigma_{\text{min}}$ to our observations for minimal level of augmentation.



From a broader perspective, conditioning augmentation addresses a well-understood issue in 
imitation learning with autoregressive-style methods, where at test-time the model must condition on samples that it produced earlier. Throughout a roll-out, the distribution of these samples may shift so that they appear out-of-distribution relative to the training data. One solution to this problem allows the model to learn from its own mistakes using a differentiable simulator~\cite{scibior2021imagining}. In the diffusion model context, though, this requires sampling from the reverse process which is expensive. We denote this type of augmentation as reverse process conditioning augmentation, where the noise originates from the model's prediction. Existing work~\cite{ho2022cascaded} on cascaded diffusion models has achieved comparable performance through both reverse process conditioning augmentation and forward process conditioning augmentation for high-resolution image generation conditioned on a low-resolution image. Therefore, we opt for the more efficient forward process conditioning augmentation approach.

\section{Experiments}

\begin{figure*}[h]
     \centering
     \begin{subfigure}[b]{0.22\textwidth}
        \centering
        \includegraphics[width=1\linewidth,keepaspectratio]{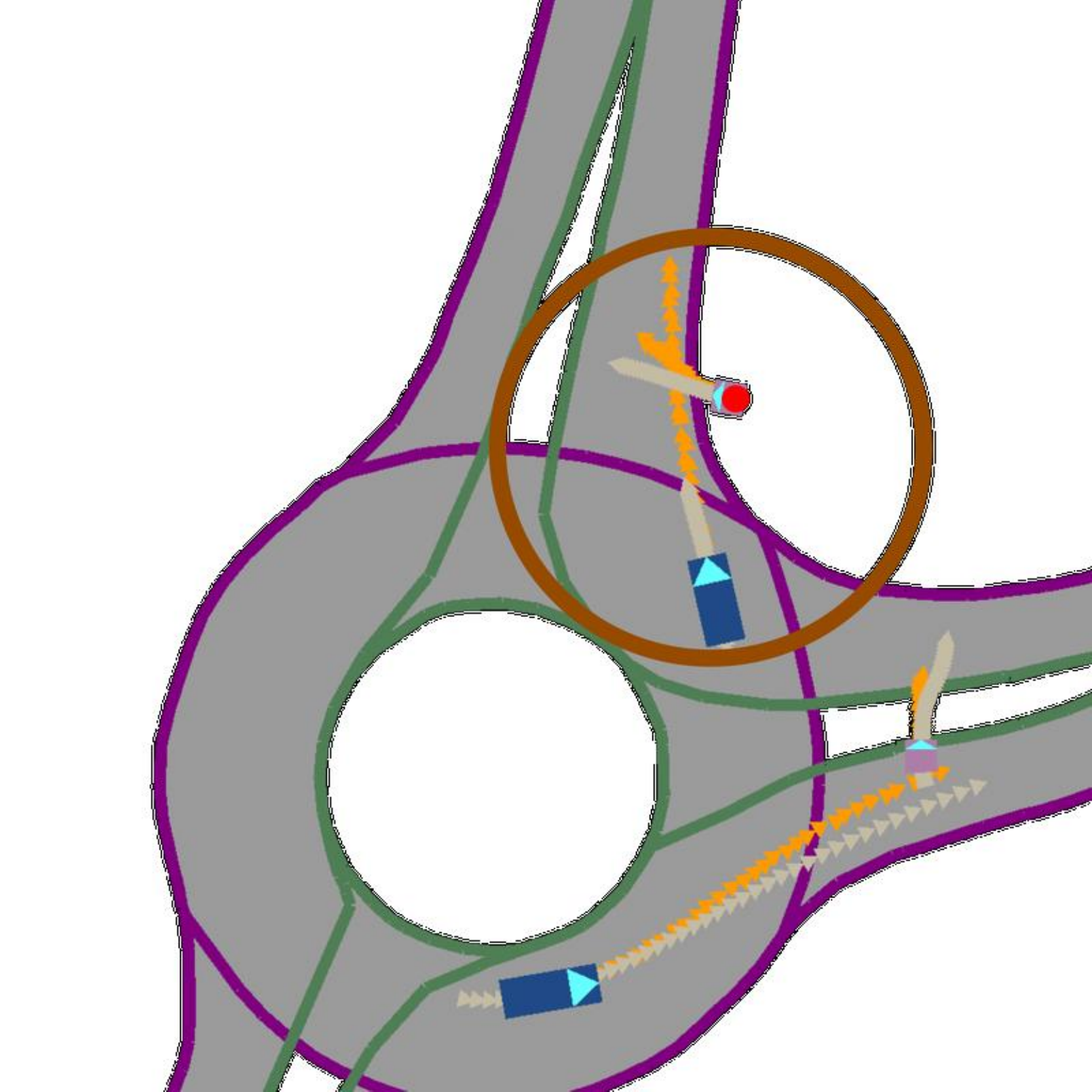}
    \end{subfigure}
    \begin{subfigure}[b]{0.22\textwidth}
        \centering
        \includegraphics[width=1\linewidth,keepaspectratio]{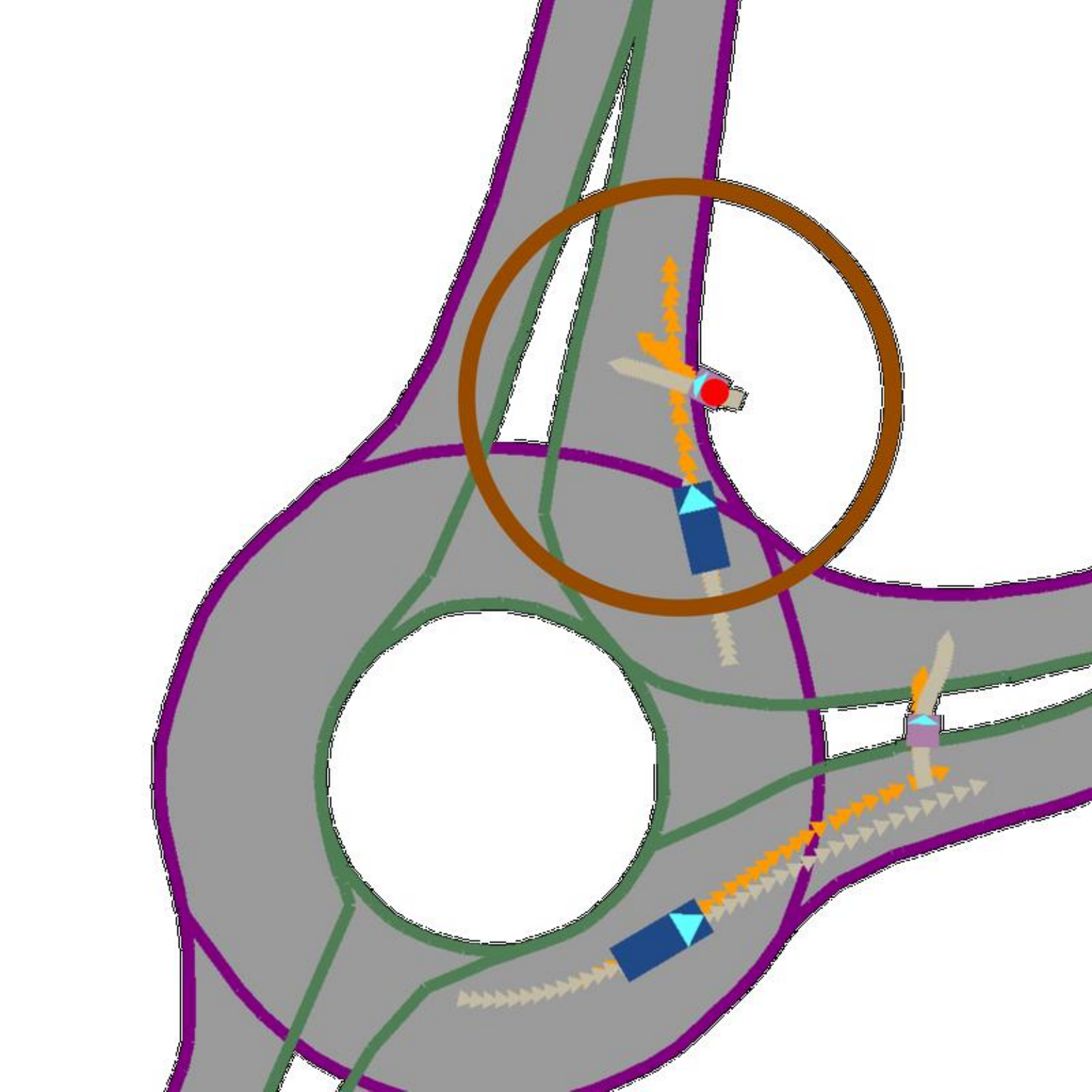}
    \end{subfigure}
    \begin{subfigure}[b]{0.22\textwidth}
        \centering
        \includegraphics[width=1\linewidth,keepaspectratio]{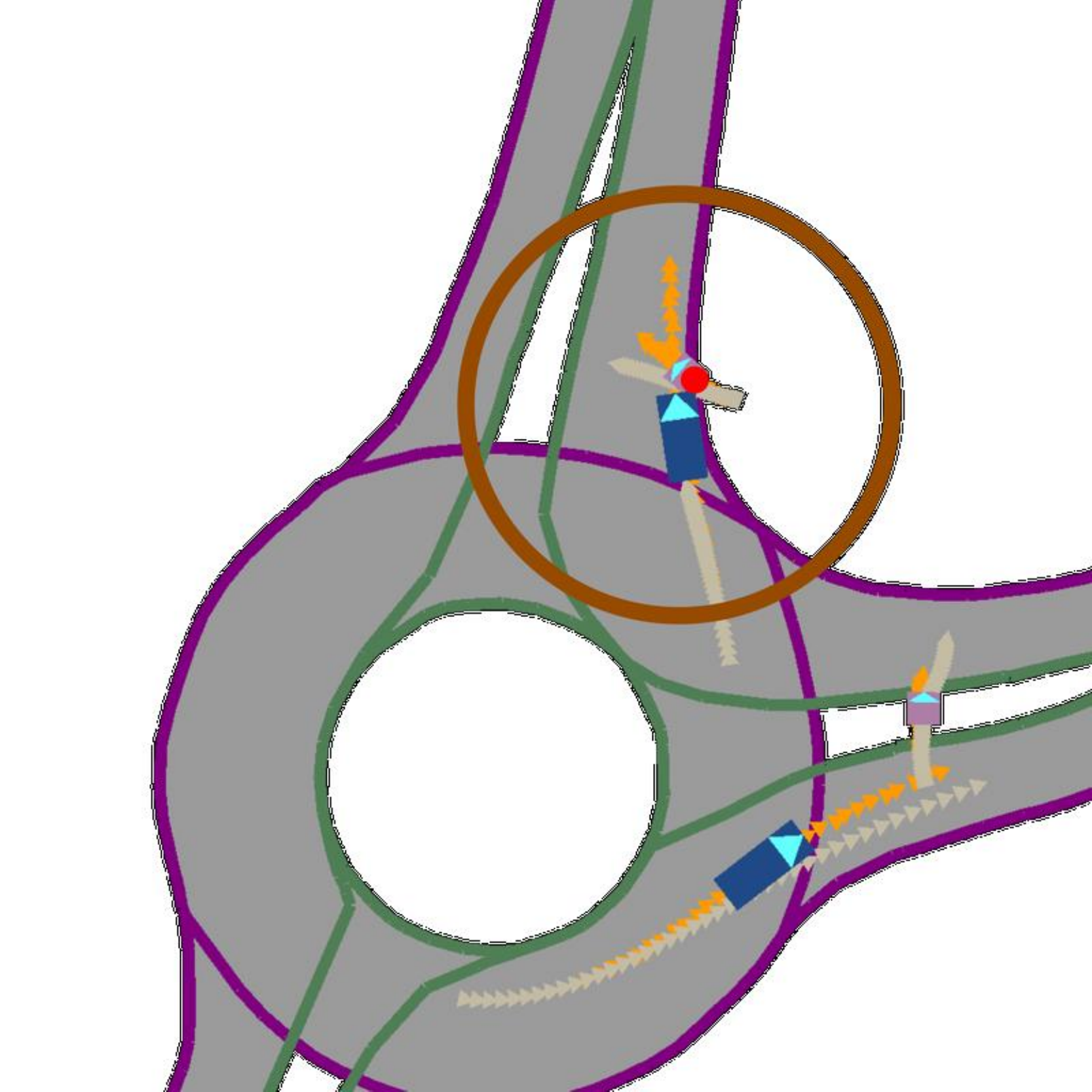}
    \end{subfigure}
        \begin{subfigure}[b]{0.22\textwidth}
        \centering
        \includegraphics[width=1\linewidth,keepaspectratio]{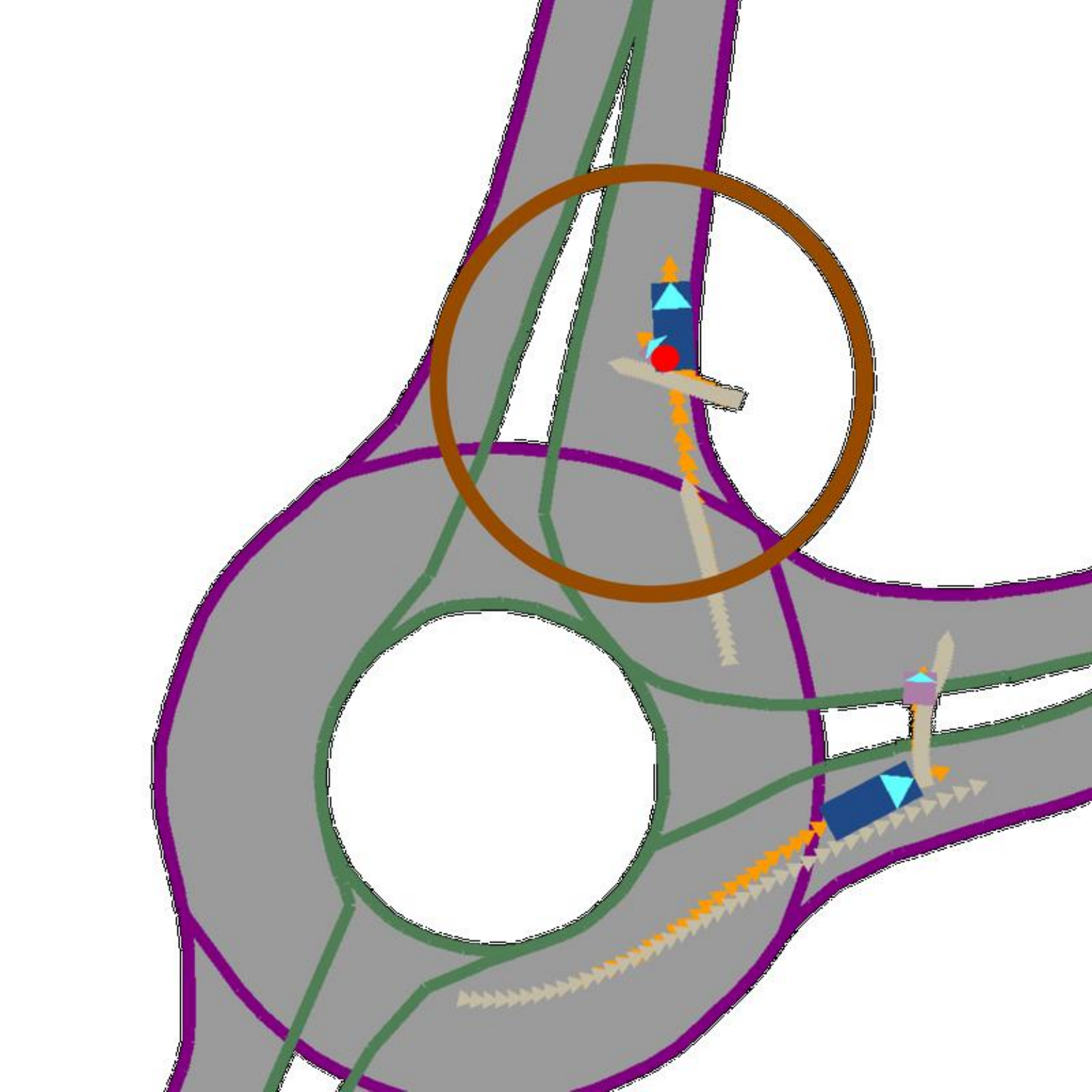}
    \end{subfigure}
        \begin{subfigure}[b]{0.22\textwidth}
        \centering
        \includegraphics[width=1\linewidth,keepaspectratio]{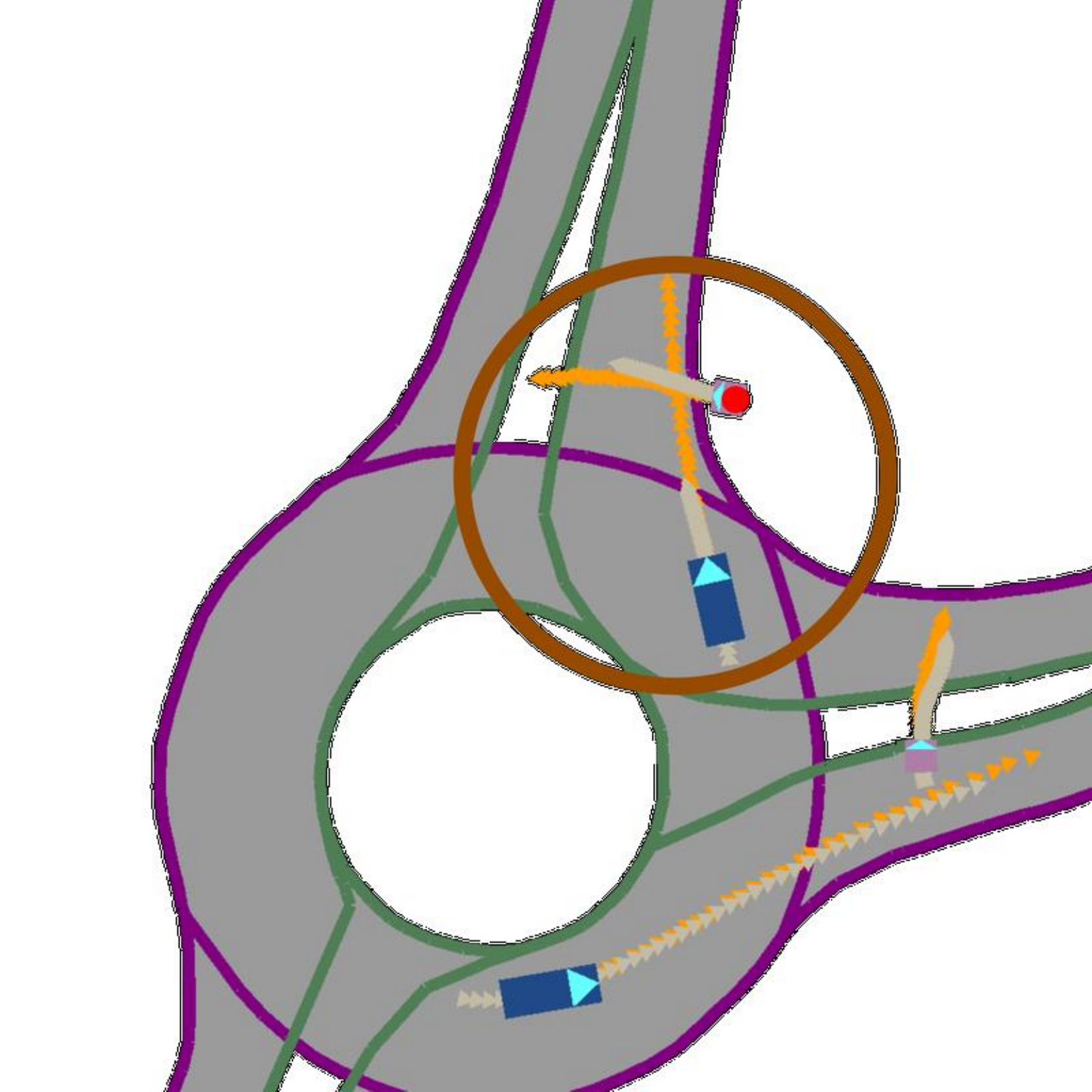}
    \end{subfigure}
            \begin{subfigure}[b]{0.22\textwidth}
        \centering
        \includegraphics[width=1\linewidth,keepaspectratio]{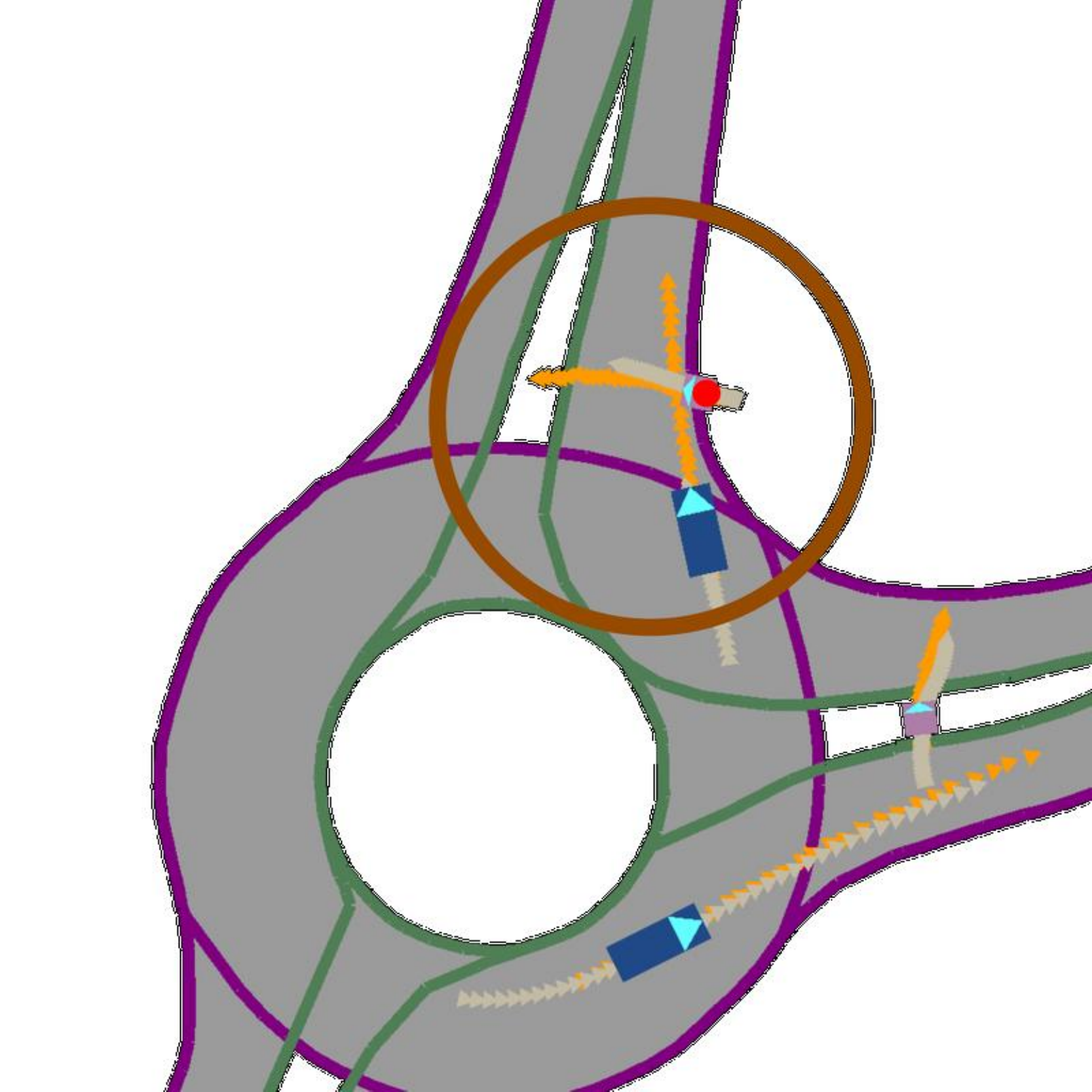}
    \end{subfigure}
        \begin{subfigure}[b]{0.22\textwidth}
        \centering
        \includegraphics[width=1\linewidth,keepaspectratio]{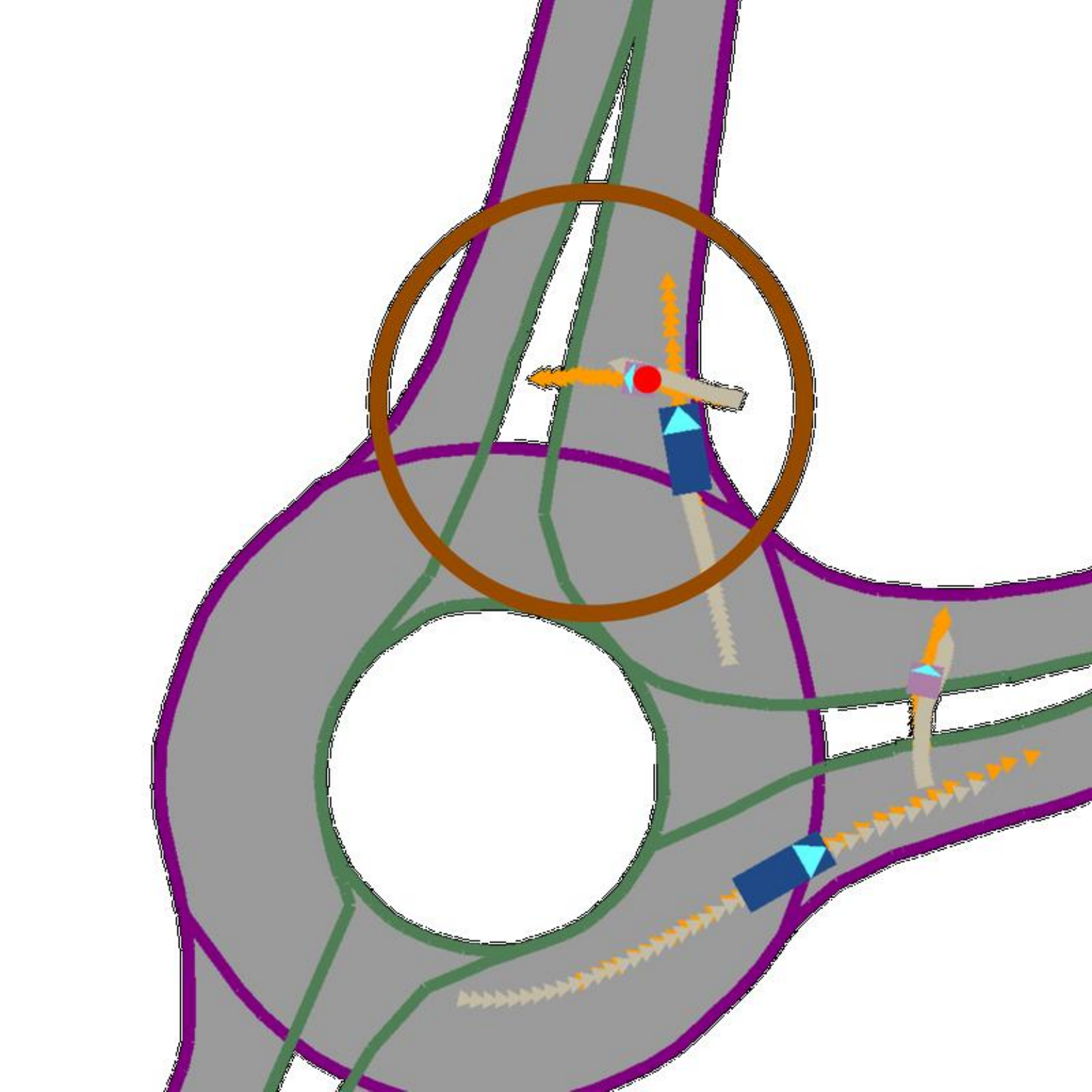}
    \end{subfigure}
        \begin{subfigure}[b]{0.22\textwidth}
        \centering
        \includegraphics[width=1\linewidth,keepaspectratio]{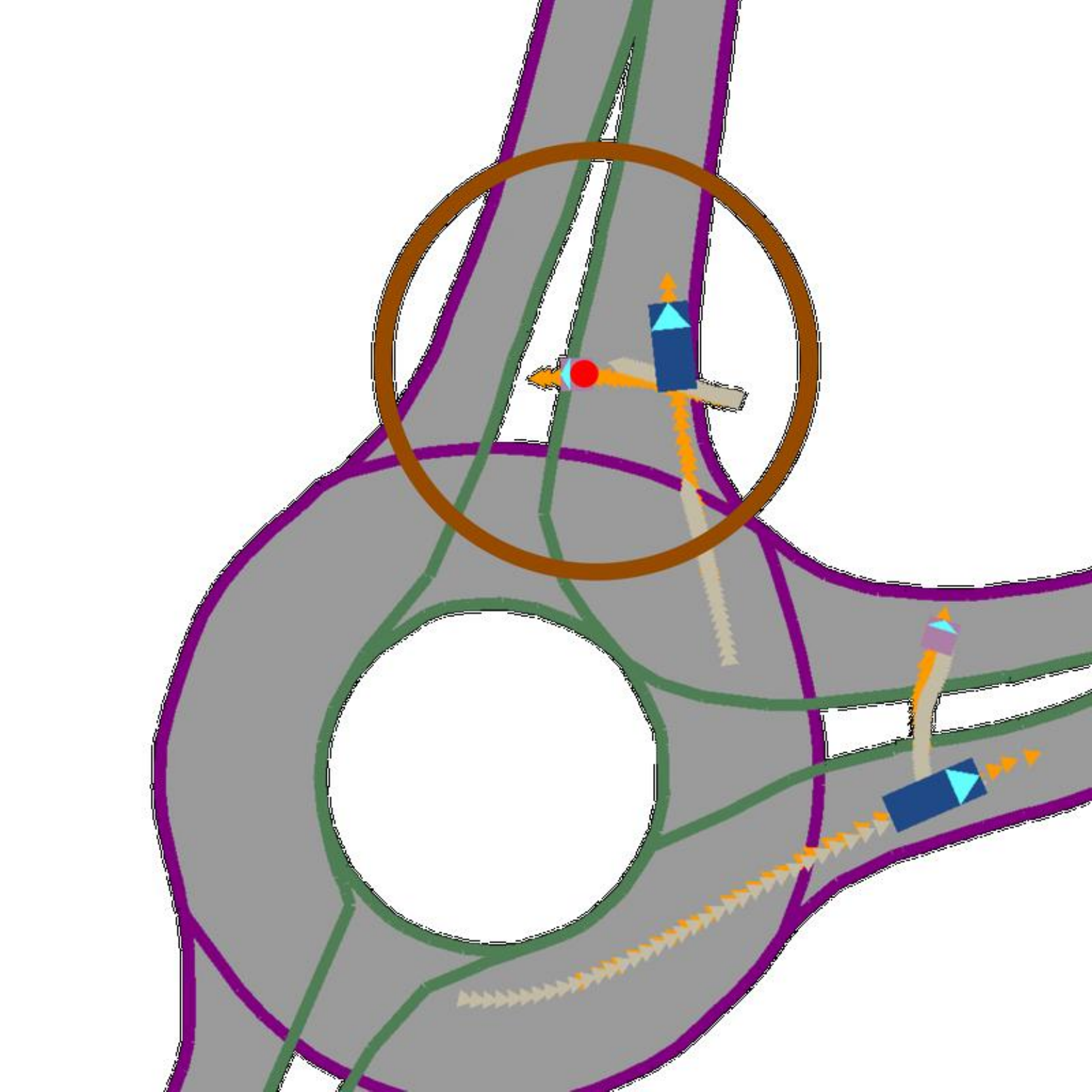}

    \end{subfigure}

    \caption{From Top to Bottom row, AR, \method{}-20. By looking ahead of the subsequent step, the pedestrian marked with a red dot controlled by \method{}-20 planner avoided colliding with the vehicle. Brown circles highlight the interaction region. Grey trajectories denote replay logs and orange trajectories are the full predicted future. This  example demonstrates that \method{}-20, with a longer planning horizon compared to AR can anticipate and mitigate interactions with other agents effectively. }
    \label{fig:dp}
\end{figure*}


We evaluate our rolling ahead scene generation model (\method{}) on the INTERACTION dataset~\cite{interactiondataset}, which contains 16.5 hours of driving records across 11 locations. Our baselines include an autoregressive diffusion model (AR), which takes observations of length 10 and predicts the next step future for all agents. Another baseline, DJINN, is a scene generation model that takes 10 observations and jointly predicts the next 30 steps at 10Hz for all agents in a one-shot manner. We have also trained a version of DJINN that predicts 10 future steps ahead jointly for all agents (DJINN-10).  As our \method{} model with window size 20 (\method{}-20) predicts 10 partially denoised future steps as well, we believe DJINN-10 provides a reasonable comparison. DJINN-10 (MPC-X) is a variant of DJINN-10 that has been trained with conditioning augmentation and deployed in an MPC style, enabling us to replan after executing X steps of predictions for all agents.

We first compare \method{} with AR, DJINN, and DJINN-10 (MPC) using standard scene-level displacement metrics such as minSceneADE and minSceneFDE to demonstrate the quality of samples generated by \method{}. 
We then assess the reactivity of DJINN, DJINN-10 (MPC-1), AR and \method{} with an adversarial agent, which is not controlled by the scene generation model, by measuring the collision rates with the adversarial agent. 

\paragraph{Implementation Details}
We adopt the same transformer architecture from DJINN \cite{niedoba2024diffusion} for all of our models. We apply 0.2 conditional augmentation for AR and DJINN-10 and 0.5 for \method{}, as we found that higher conditional augmentation ratio for AR and DJINN-10 results in worse performance. We train our \method{} planner with observation length 10 and task ratio $\beta$=0.1.

\paragraph{Evaluation Metrics}

We measure the accuracy of our generated trajectories with standard displacement metrics. To measure the joint motion forecasting quality, we follow \cite{ngiam2021scene} reporting minSceneADE and minSceneFDE. Both metrics capture the minimum joint displacements error for all agents across 6 joint traffic scenario samples. To measure per-agent motion forecasting performance, we report the miss rate; the rate of agents where none of the six predicted trajectories have a final displacement error less than 2 meters. To measure the reactivity of each model, we report the collision rate, the number of collisions divided by the total number of simulated scenarios.


\paragraph{Motion Forecasting}
We compare \method{} with AR and DJINN-10 on the motion forecasting task using the validation set of the INTERACTION dataset \cite{interactiondataset}. We generate three seconds of driving behavior at 10 hertz, conditioned on one second of observations. We consider the performance of DJINN as the upper bound for this task since DJINN is trained only at this fixed time horizon and is not an autoregressive model by nature. Following~\cite{niedoba2024diffusion}, displacement metrics are calculated by generating 24 samples for each scenario and fit a 6 component Gaussian mixture model to cover all future modes. DJINN-10 (MPC-1) achieves slightly better results than AR but performs worse than \method{} due to larger accumulated errors from replanning at each simulation step. 

\method{} models with window sizes of 15 (\method{}-15) and 20 (\method{}-20) achieved lower displacement metrics than AR models, as \method{} also considers the noisy future steps beyond the next immediate one. Additionally, \method{} models exhibit a slightly lower miss rate. We also observed that \method{} models with larger window sizes demonstrate better displacement metrics. Displacement metrics are one indicator of the quality of the generated samples. We show qualitatively in Figure~\ref{fig:dp} that \method{} reconstructs to ground truth trajectories marked in grey better than AR. 

\begin{table*}[ht]
    \centering
    \caption{Comparison of displacement metrics and collision rate. }
    \begin{tabular}{llrrr}
        \toprule
        \multicolumn{2}{c}{\textbf{Location}} & \multicolumn{3}{c}{\textbf{All}} \\
        \cmidrule(r){1-2} \cmidrule(r){3-5}
        \textbf{Model} & \textbf{Type} & \textbf{minSceneADE} & \textbf{minSceneFDE} & \textbf{Miss Rate}  \\
                 {DJINN} & (One shot) & 0.388& 1.004& 0.049  \\  \hline
                     
                 {DJINN-10} & (MPC-1)& 0.692& 1.675& 0.166 \\
        AR & & 0.695 & 1.670  & 0.168  \\
        \method{}-15 & & {0.673} & {1.596} & {0.160}\\
        \method{}-20 & & \textbf{0.654} & \textbf{1.553} & \textbf{0.142} \\
        \bottomrule
    \end{tabular}%
    \label{tab:perf}
\end{table*}

\paragraph{Reactivity}

\begin{table}[ht]
    \centering
    \caption{Performance with an adversarial ego agent.}
    \begin{tabular}{lrr}
        \toprule
        \textbf{Model} & \textbf{Collision Rate} &\textbf{Prediction time} (min) \\
        \midrule
        DJINN     & 0.052 & 0.07 \\\hline
        \makecell{DJINN-10 \\ (MPC-1)} & \textbf{0.014} & 0.69 \\
        AR       & 0.016 & 0.68 \\
        {\method{}-15} & {0.019} &  {0.34} \\
        \method{}-20  & 0.024 & \textbf{0.20}\\
        \bottomrule
    \end{tabular}
    \label{tab:collision_scores}
\end{table}

The \method{} models efficiently roll out long scenarios by partially denoising future states.  However, this limits its ability to adapt to perturbations, such as an agent controlled by a different motion planner while being observed by our model.  This is a typical setup in closed-loop simulation. To evaluate this, we evaluate the \method{} models' adaptation to an adversarial agent. We select one agent per scene and control it using its replay log, slowing it down to reach only half its trajectory by the end of the simulation. The simulation runs for 40 time steps at 10 Hz, given initial 10-step observations, which makes the performance of DJINN one-shot a lower bound since it is blind to the adversarial agent during the simulation.

In total, we select 1,440 scenes from the INTERACTION validation set, focusing on the top six locations with the largest number of scenarios. Scenes with a low number of participants are filtered out, as agents in these scenarios are less likely to interact. We take three samples per scenario and for each model, then report the average collision rate in Table~\ref{tab:perf}. In addition, we reported the prediction time for a single sample on a RTX2080Ti GPU in Table~\ref{tab:collision_scores} for each model to highlight the efficiency of our \method{} models. 

DJINN-10 (MPC-1) achieves the lowest collision rate compared to other models while having the longest total prediction time.  AR reduces the collision rate by 3x compared to the lower bound (DJINN). DJINN-10 (MPC-1) achieving better results than AR aligns with our expectations, as DJINN-10 (MPC-1)can look ahead ten steps into the future, whereas AR predicts only one step ahead. Our proposed \method{}-15 performs close to AR which reduces the collision rate over 2.5x compared to DJINN while having half of the prediction time compared to AR and DJINN-10 (MPC-1).

As an ablation, we measured the collision rate for \method{}-20 in the reactivity experiment.  We observed that increasing the window size can reduce the prediction time further while having a slightly higher collision rate. Figure S1 in the Supplementary Materials shows an example where \method{}-20 failing to react, while \method{}-15 avoids a collision in the same scenario. This feature provides practitioners with the flexibility to decide whether reactivity or computational efficiency is more important for their simulation needs.

\paragraph{Ablation on conditioning augmentation}
We show the significance of conditioning augmentation by measuring the displacement metrics for two \method{} models trained with same configuration but one without conditioning augmentation in Table~\ref{tab:abla}.  We can see the displacement errors increased significantly without conditioning augmentation.


\begin{table}[th!]
    \centering
    \caption{Ablation on conditioning argumentation (CA) across six locations from INTERACTION dataset.}
    \begin{tabular}{l|ll}
    \textbf{Metrics} & \method{} w/o CA &  \method{} w CA \\ \hline
    minSceneADE &  0.930 & \textbf{0.663}\\ 
    minSceneFDE & 2.197 & \textbf{1.579}\\
    ego\_minADE & 0.693 & \textbf{0.475}
    \end{tabular}
    \label{tab:abla}
\end{table}

\section{Conclusion}

In conclusion, we have proposed a rolling diffusion-based traffic scene planning framework that strikes a beneficial compromise between reactivity and computational efficiency. We believe this work addresses a gap in the community by enabling the autoregressive generation of traffic scenarios for all agents jointly, and it offers insights into the crucial role of conditioning augmentation techniques. For future work, we aim to explore test-time conditioning with this model and seek to enhance model performance through flexible conditioning on past observations~\cite{harvey2022flexible}.

\section{Acknowledgment}
We acknowledge the support of the Natural Sciences and Engineering Research Council of Canada (NSERC), the Canada CIFAR AI Chairs Program, Inverted AI, MITACS, and Google. This research was enabled in part by technical support and computational resources provided by the Digital Research Alliance of Canada Compute Canada (alliancecan.ca), the Advanced Research Computing at the University of British Columbia (arc.ubc.ca), and Amazon.


\bibliography{aaai25}
\clearpage

\section{Supplementary Materials}

\subsection{Introduction}
We provide an extended discussion on related work regarding autoregressive diffusion models. We also detail the computational resources and datasets used in our experiments. Furthermore, we present additional results on accumulation errors caused by replanning for DJINN-10, as well as  visualizations showcasing the reactivity of the \method{} model with varying window sizes.

\subsection{Additional Related Work}
Autoregressive Diffusion Models (ARDM)~\cite{hoogeboom2021autoregressive} introduce an order-agnostic autoregressive diffusion model that combines an order-agnostic autoregressive model~\cite{uria2014deep} with a discrete diffusion model~\cite{austin2021structured}. The order-agnostic nature of this model eliminates the need for generating subsequent predictions in a specific order, thereby enabling faster prediction times through parallel sampling. Additionally, relaxing the causal assumption leads to a more efficient per-time-step loss function during training. However, such a model is not suitable for our application due to the sequential nature of traffic simulation. AMD~\cite{han2024amd} proposes an auto-regressive motion generation approach for human motion given a text prompt, but unlike the Rolling Diffusion Model, it denoises one clean motion sample at a time, which is slow at prediction time. The Rolling Diffusion Model (RDM)~\cite{ruhe2024rolling} proposes a sliding window approach targeted at long video generation but does not specifically study its application in a multi-agent system, particularly for closed-loop traffic simulation. We investigate the level of reactivity when applying rolling diffusion models as a traffic scene planner.

\subsection{Compute resources}

We run all our experiments on four NVIDIA V100 GPUs hosted by a cloud provider. We trained our \method{} models for 9 days, and so 36 GPU-days. AR and DJINN were also trained for 36 GPU-days. In total, including preliminary runs and ablations, we estimate that the project required roughly 300 GPU-days.

\subsection{Dataset}

We experiment with the INTERACTION dataset~\cite{interactiondataset} which is available for non-commercial use following the guidelines at \url{https://interaction-dataset.com/}.

\subsection{Accumulation Errors for DJINN-10}

We observe that DJINN-10, even when trained with conditional augmentation, still experiences accumulation errors caused by autoregressive replanning. In Table~\ref{tab:abla-mpc}, we compare the displacement error of DJINN-10 at a replanning rate of 10 Hz (MPC-1) and 2 Hz (MPC-5) for a prediction horizon of 40 with an observation length of 10. DJINN-10 (MPC-1) exhibits significantly higher displacement error. In contrast, \method{} utilizes a sliding window approach with decreasing SNR ratio within the window. Denoising for the next simulation step does not start from Gaussian noise, resulting in lower accumulation errors compared to DJINN-10 (MPC-1).

\begin{table}[th!]
    \centering
    \caption{Accumulation Errors caused by replanning for DJINN-10}
    \begin{tabular}{l|ll}
    \textbf{Metrics} & DJINN-10 (MPC-1) &  DJINN-10 (MPC-5) \\ \hline
    minSceneADE &  0.692 & \textbf{0.583}\\ 
    minSceneFDE & 1.675 & \textbf{1.351}\\
    Miss Rate & 0.166 & \textbf{0.091}
    \end{tabular}
    \label{tab:abla-mpc}
\end{table}

\subsection{Additional Visualizations}
\begin{figure*}[t!]
     \centering
     \begin{subfigure}[b]{0.225\textwidth}
        \centering
        \includegraphics[width=1\linewidth,keepaspectratio]{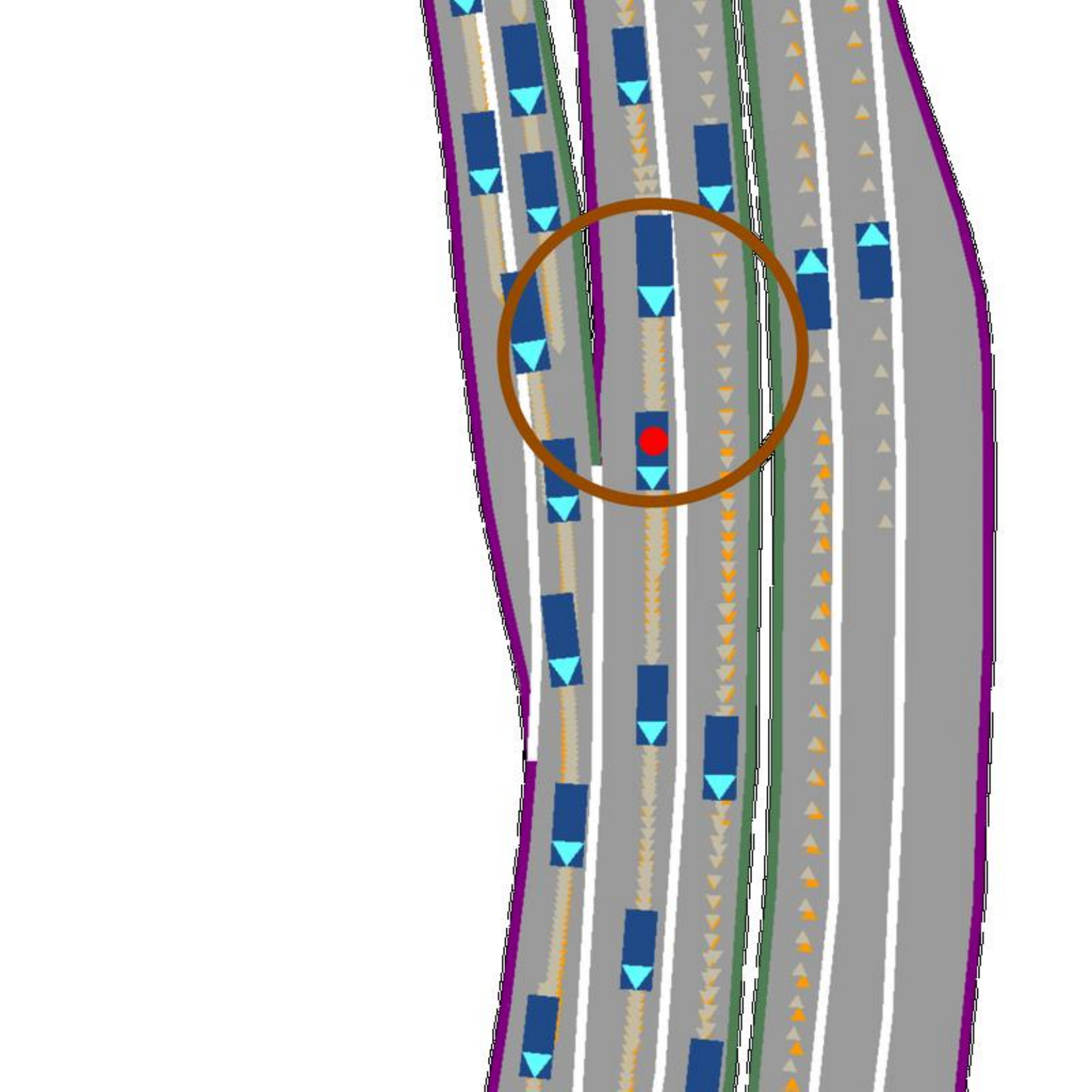}
    \end{subfigure}
    \begin{subfigure}[b]{0.225\textwidth}
        \centering
        \includegraphics[width=1\linewidth,keepaspectratio]{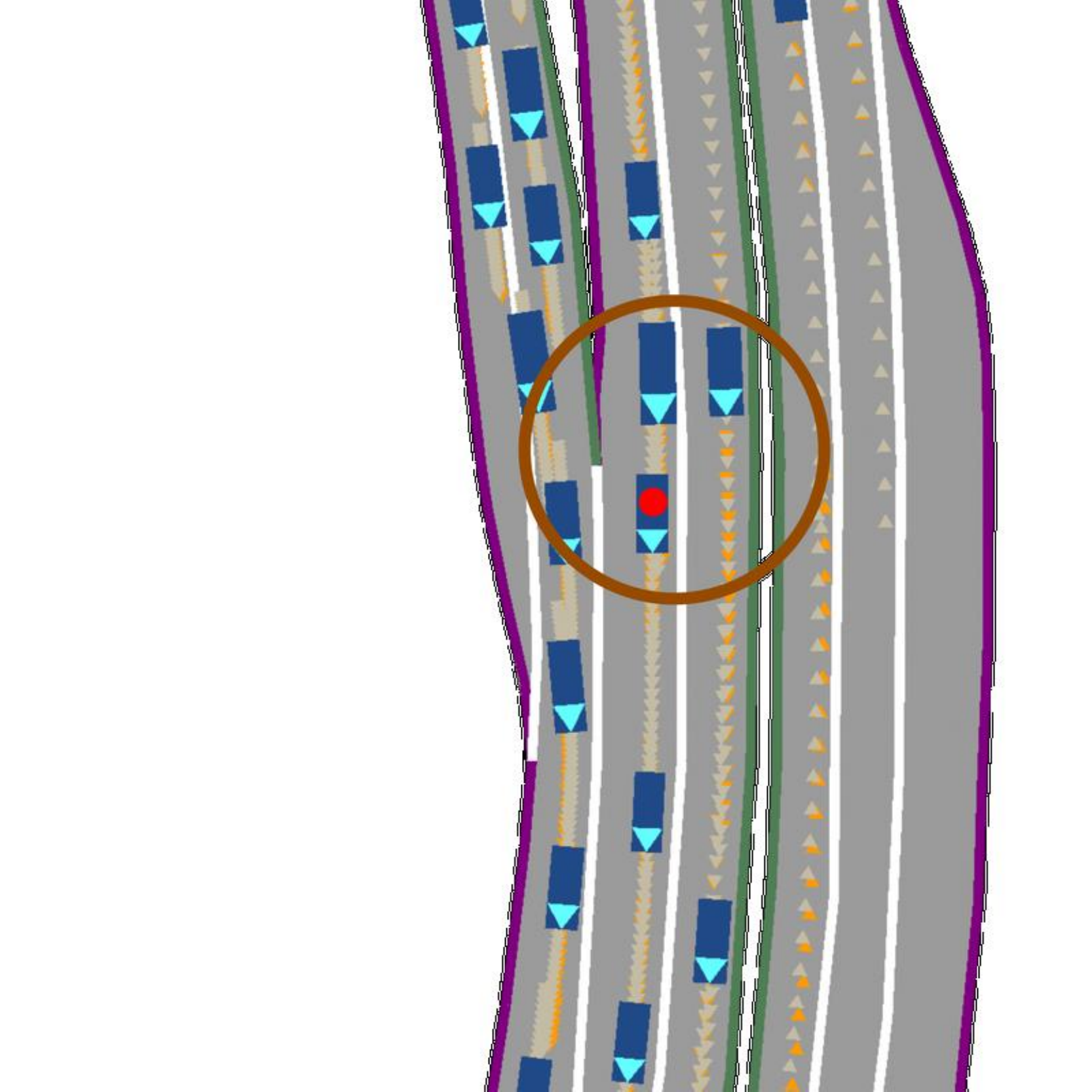}
    \end{subfigure}
    \begin{subfigure}[b]{0.225\textwidth}
        \centering
        \includegraphics[width=1\linewidth,keepaspectratio]{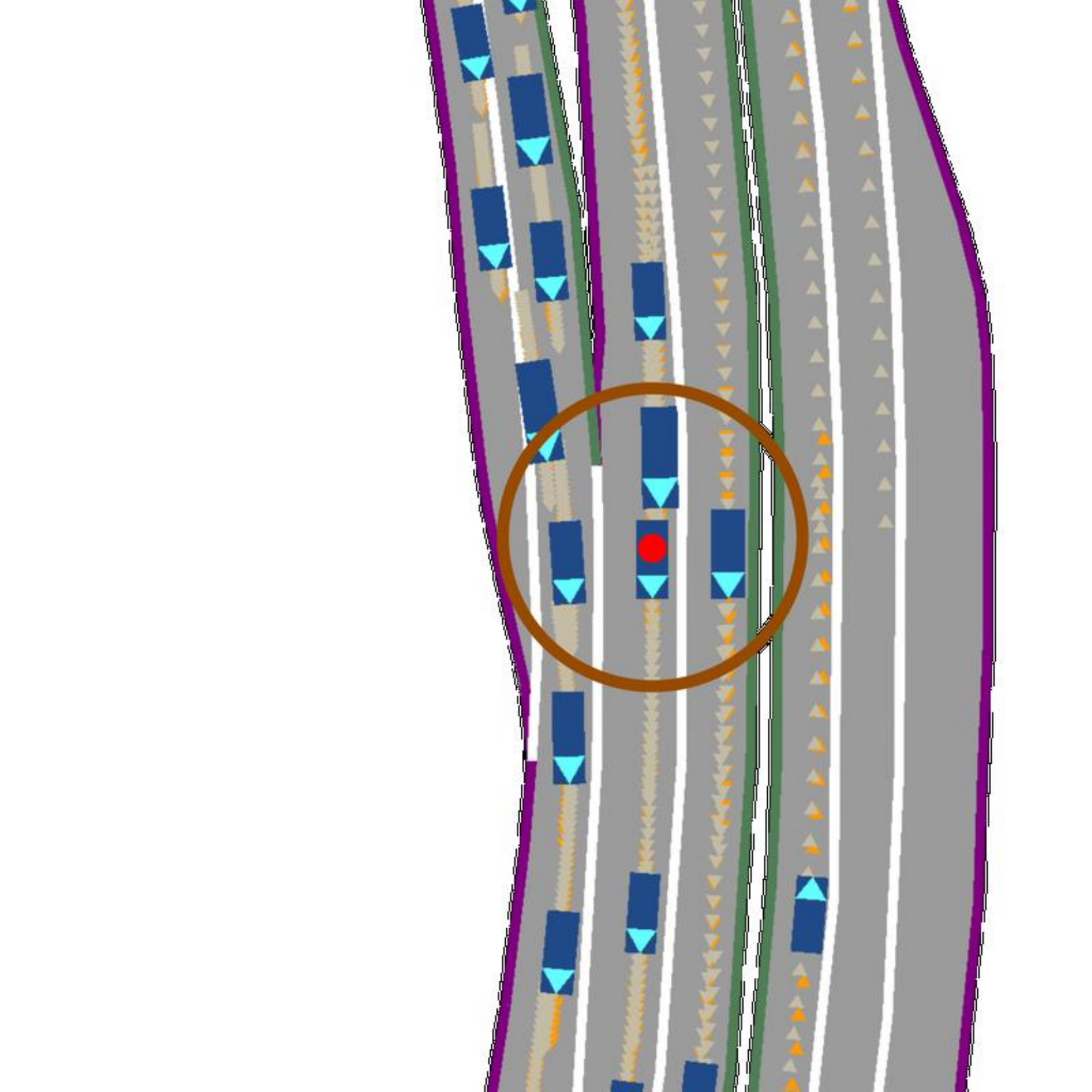}
    \end{subfigure}
        \begin{subfigure}[b]{0.225\textwidth}
        \centering
        \includegraphics[width=1\linewidth,keepaspectratio]{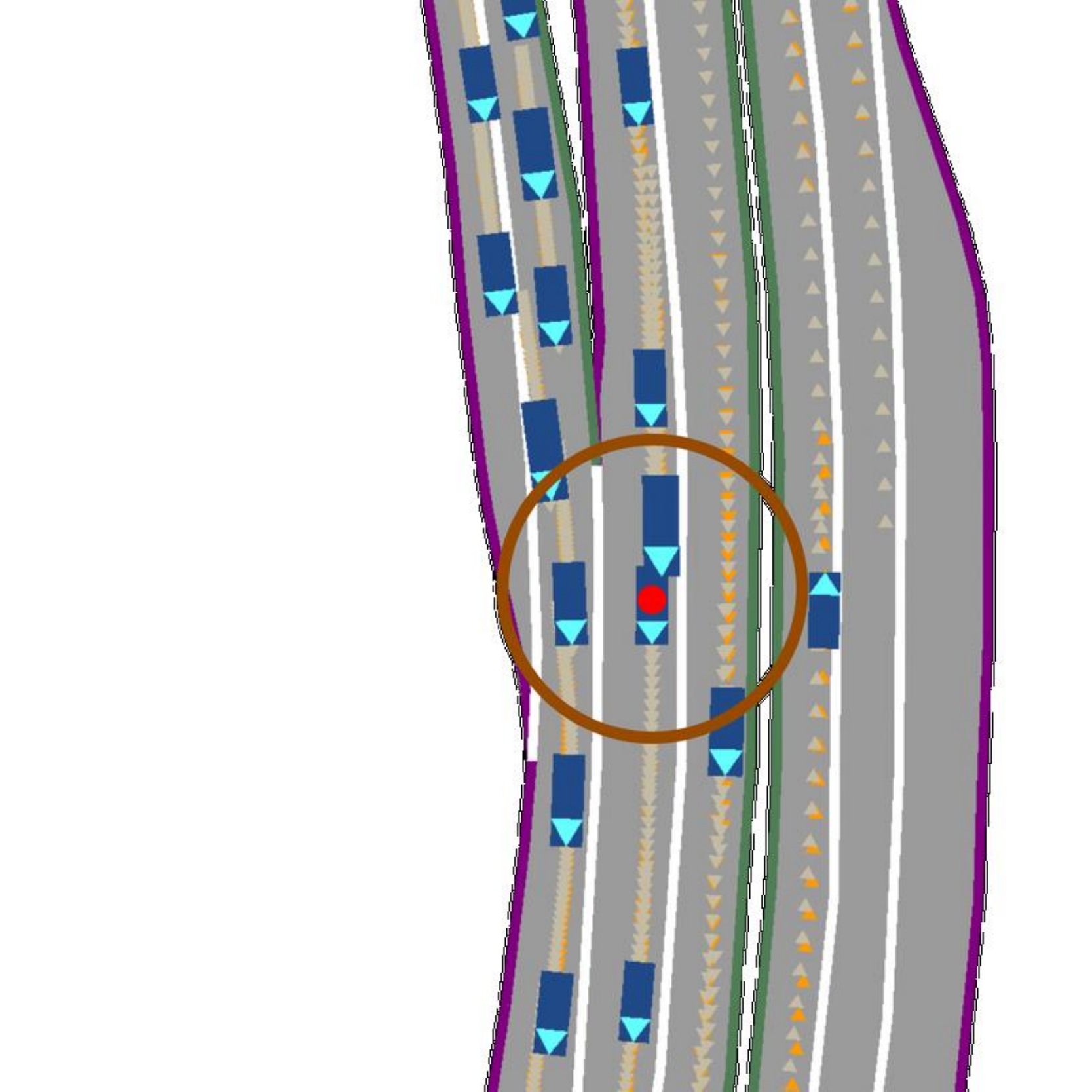}
    \end{subfigure}
        \begin{subfigure}[b]{0.225\textwidth}
        \centering
        \includegraphics[width=1\linewidth,keepaspectratio]{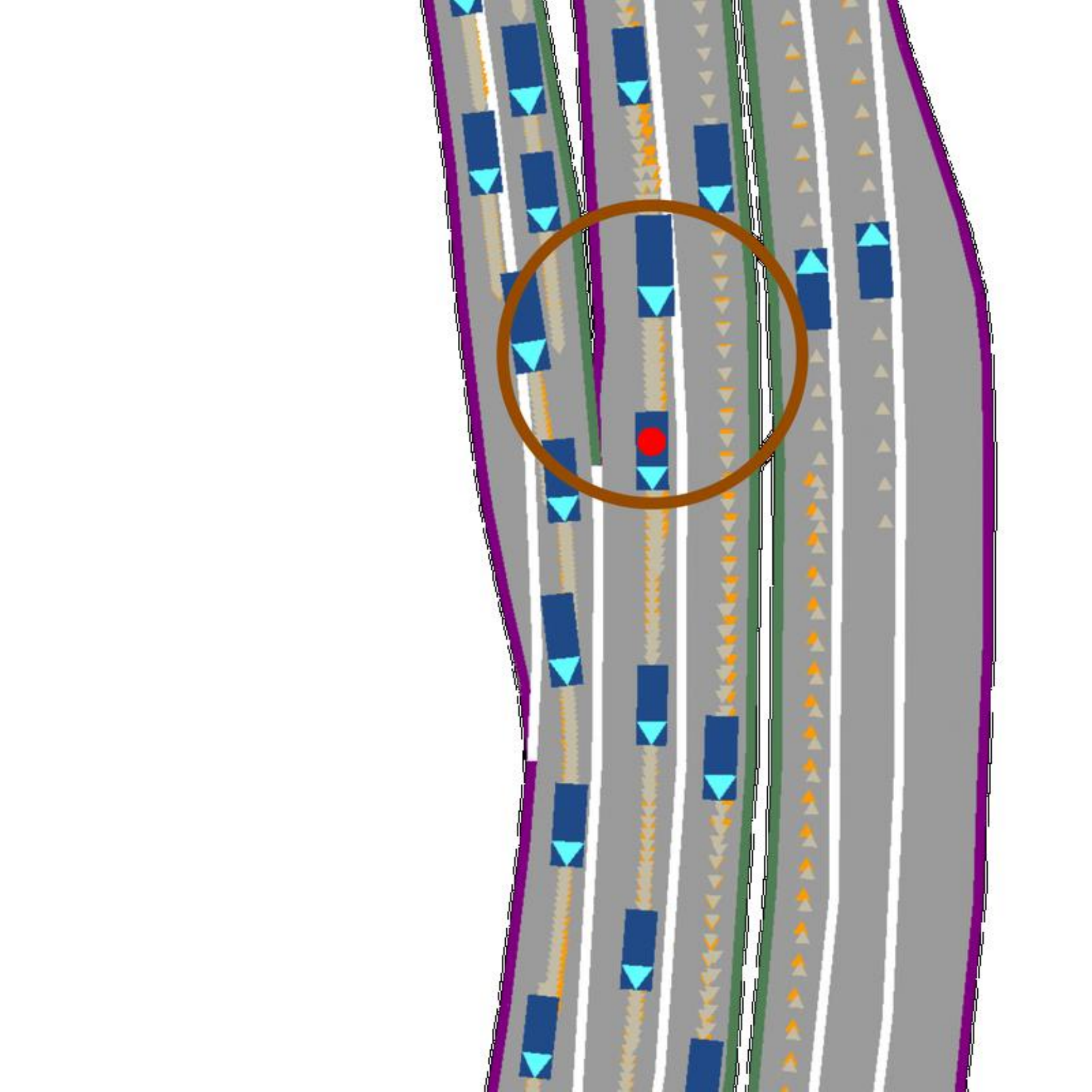}
    \end{subfigure}
            \begin{subfigure}[b]{0.225\textwidth}
        \centering
        \includegraphics[width=1\linewidth,keepaspectratio]{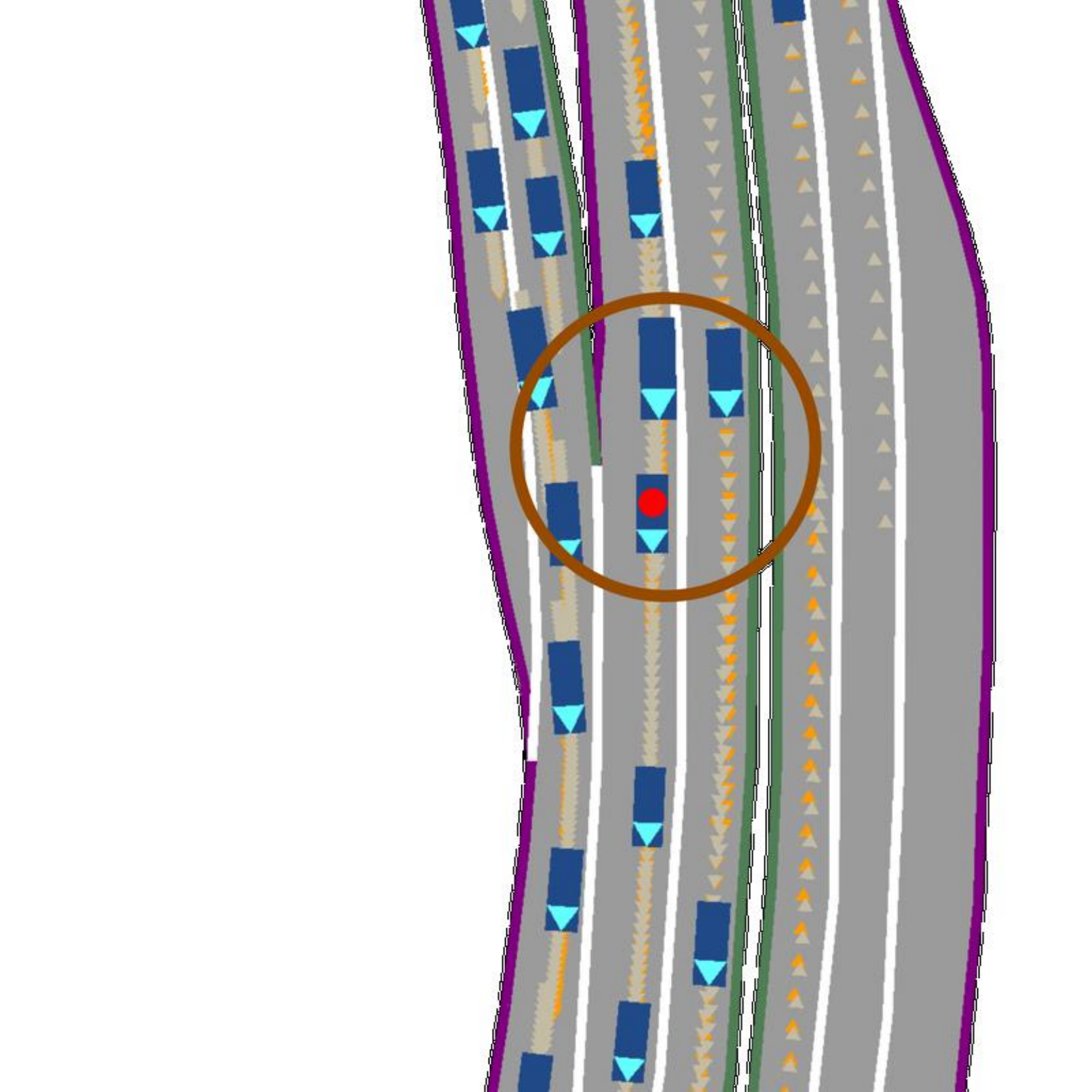}
    \end{subfigure}
        \begin{subfigure}[b]{0.225\textwidth}
        \centering
        \includegraphics[width=1\linewidth,keepaspectratio]{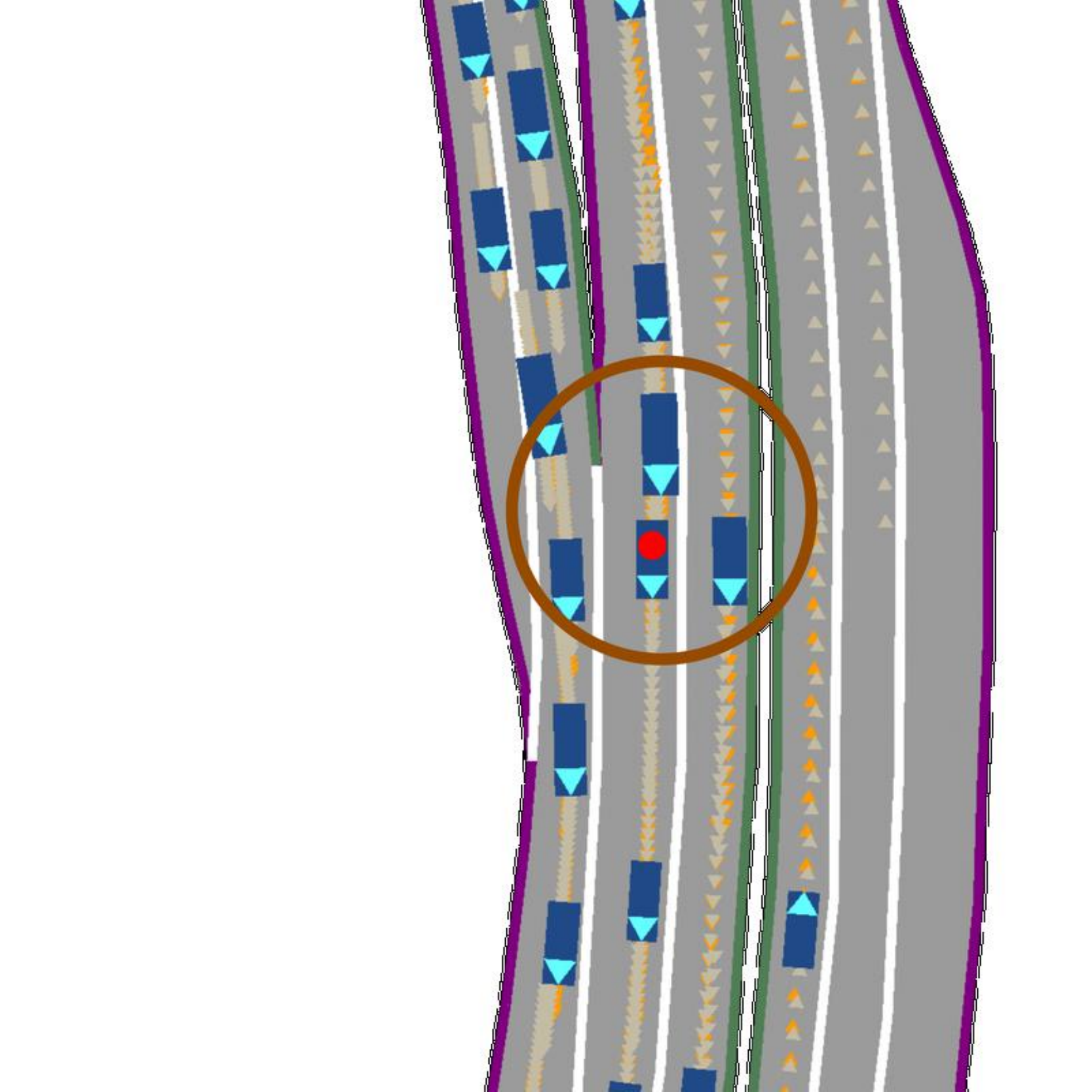}
    \end{subfigure}
        \begin{subfigure}[b]{0.225\textwidth}
        \centering
        \includegraphics[width=1\linewidth,keepaspectratio]{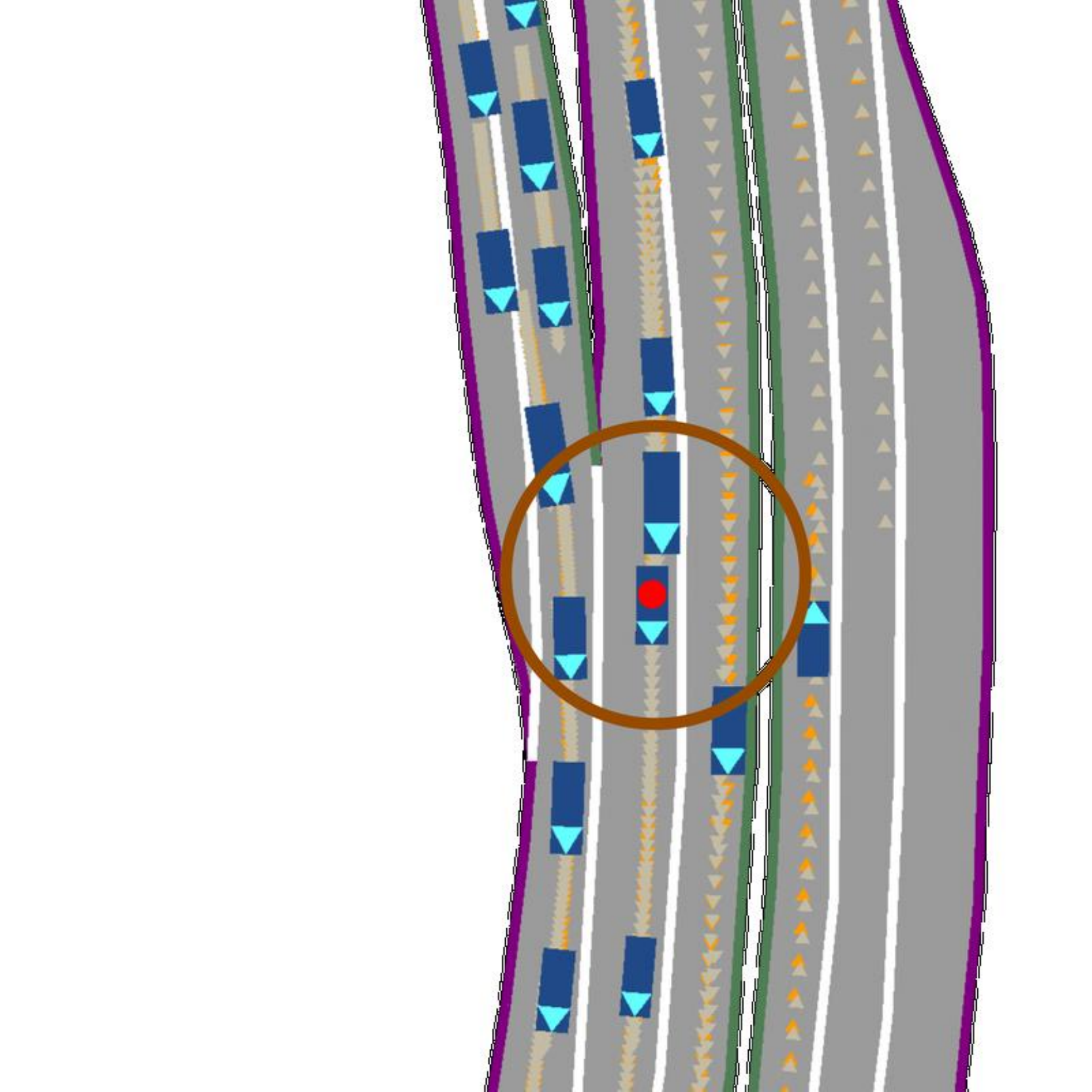}

    \end{subfigure}

    \caption{From top to bottom row: \method{}-20, \method{}-15. The adversarial agent, marked with a red dot, follows its replay log and slows down to reach only half its trajectory by the end of the simulation. Brown circles highlight the interaction region. \method{}-15 achieves better reactivity than \method{}-20, as reducing the window size causes the model to denoise the next element from a lower signal-to-noise ratio (SNR), which provides the model with greater flexibility to adjust to the adversarial agent.}
    \label{fig:traj-v-aug}
\end{figure*}

In Figure \ref{fig:traj-v-aug}, we demonstrate that the flexibility of our \method{} model by adjusting the sliding window size.

\end{document}